\documentclass[journal]{IEEEtran}

\usepackage{times}
\usepackage{epsfig}
\usepackage{graphicx}
\usepackage{amsmath}
\usepackage{amssymb}
\usepackage{algorithm} 
\usepackage{algorithmic} 
\usepackage{multirow} 
\usepackage{xcolor}
\usepackage{paralist}
\usepackage{pifont}
\usepackage{array}
\usepackage{cite}

\ifCLASSINFOpdf
\else
\fi

\usepackage{xcolor}

\begin{document}
\title{Multiscale Spatio-Temporal  Graph  Neural Networks for 3D Skeleton-Based Motion Prediction}

\author{Maosen~Li,~\IEEEmembership{Student Member,~IEEE,}
        Siheng~Chen,~\IEEEmembership{Member,~IEEE,}
        Yangheng~Zhao,
        Ya~Zhang,~\IEEEmembership{Member,~IEEE,}
        Yanfeng~Wang,
        and~Qi~Tian,~\IEEEmembership{Fellow,~IEEE}
\thanks{M. Li, Y. Zhao, Y. Zhang and Y. Wang are with the Cooperative Medianet Innovation Center, Shanghai Jiao Tong University, Shanghai, 200240 China. S. Chen is with Mitsubishi Electric Research Laboratories, Cambridge, MA, USA. Q. Tian is with Shanghai Jiao Tong University, Shanghai, 200240 China.}
\thanks{Manuscript received April 19, 2005; revised August 26, 2015.}}

\markboth{IEEE TRANSACTIONS ON IMAGE PROCESSING,}
{Li \MakeLowercase{\textit{et al.}}: Multiscale Spatio-Temporal  Graph  Neural Networks for 3D Skeleton-Based Motion Prediction}
\maketitle

\begin{abstract}
We propose a multiscale spatio-temporal graph neural network (MST-GNN) to predict the future 3D skeleton-based human poses in an action-category-agnostic manner. The core of MST-GNN is a multiscale spatio-temporal graph that explicitly models the relations in motions at various spatial and temporal scales. Different from many previous hierarchical structures, our multiscale spatio-temporal graph is built in a~\emph{data-adaptive fashion}, which captures nonphysical, yet motion-based relations. The key module of MST-GNN is a multiscale spatio-temporal graph computational unit (MST-GCU) based on the trainable graph structure. MST-GCU embeds underlying features at individual scales and then fuses features across scales to obtain a comprehensive representation. The overall architecture of MST-GNN follows an encoder-decoder framework, where the encoder consists of a sequence of MST-GCUs to learn the spatial and temporal features of motions, and the decoder uses a graph-based attention gate recurrent unit (GA-GRU) to generate future poses. Extensive experiments are conducted to show that the proposed MST-GNN outperforms state-of-the-art methods in both short and long-term motion prediction on the datasets of Human 3.6M, CMU Mocap and 3DPW, where MST-GNN outperforms previous works by $5.33\%$ and {$3.67\%$} of mean angle errors in average for short-term and long-term prediction on Human 3.6M, and by {$11.84\%$} and {$4.71\%$} of mean angle errors for short-term and long-term prediction on CMU Mocap, and by {$1.13\%$} of mean angle errors on 3DPW in average, respectively. We further investigate the learned multiscale graphs for interpretability.
\end{abstract}


\IEEEpeerreviewmaketitle


\section{Introduction}
\IEEEPARstart{3}{D} skeleton-based human motion prediction aims to forecast the future 3D poses composed of key body-joints and skeleton structures by exploiting the dynamics of the past motions, which is recently paid considerable attention~\cite{Jain_2016_CVPR, Martinez_2017_CVPR, Butepage_2017_CVPR, Gui_2018_ECCV,Walker_2017_ICCV} to potentially helping intelligence systems understand complex human behaviors and serve in many computer vision and robotics scenarios, e.g. human-computer interaction~\cite{gui-2018-110272}, autonomous driving~\cite{SihengChen_IEEE_SP}, and pedestrian tracking~\cite{Alahi_2016_CVPR, Bhattacharyya_2018_CVPR}.

\begin{figure}[t]
    \centering
    \includegraphics[width=0.9\columnwidth]{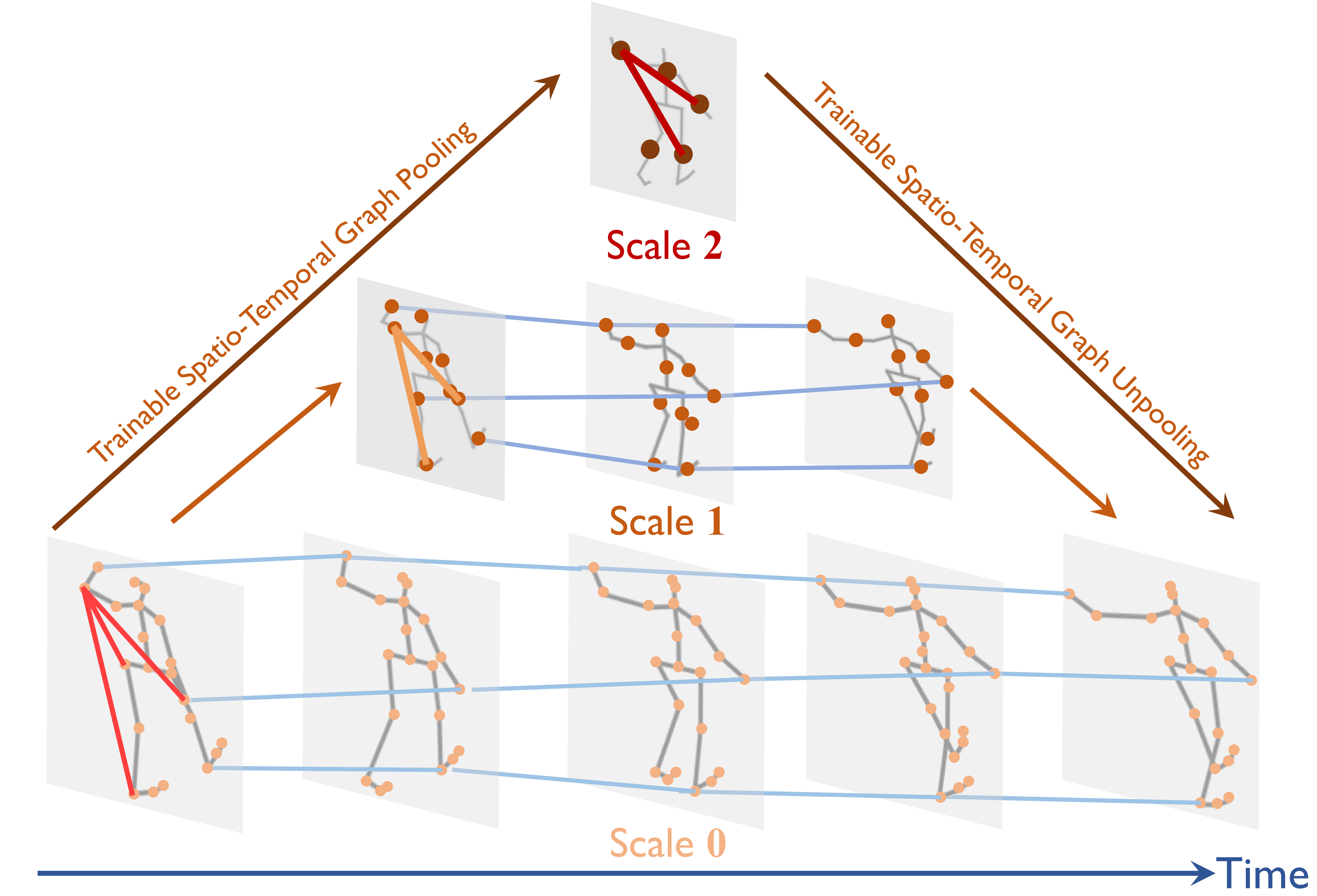}
    \caption{The proposed spatio-temporal graph structure is trainable, decomposable and multiscale. At each scale, the spatio-temporal graph can be decomposed into a trainable spatial graph and a trainable temporal graph.  Through trainable spatial graph pooling/unpooling and temporal graph pooling/unpooling, we are able to convert graphs across scales.
    }
    \vspace{-5mm}
    \label{fig:Jewel3}
\end{figure}

The main challenge of motion prediction is to handle the randomness, flexibility, and nonlinearity in the real-world human motions. Many methods have been proposed to tackle this challenge, including the conventional state-based methods~\cite{Lehrmann_2014_CVPR, NIPS2006_3078} and deep-network-based methods~\cite{Fragkiadaki_2015_ICCV, Martinez_2017_CVPR, Hsukuang18, Gui_2018_ECCV, AAAI_Guo, Gopalakrishnan_2019_CVPR, quater,WangSpatio}, which arrange the pose features into pseudo images to learn the movement patterns. However, those methods rarely exploit the spatial or temporal relations between the body-joints. The spatial relations capture the inherent pose constraints; and the temporal relations capture the inter-frame correlation to depict the continuous dynamics.

To employ the internal motion relations, recent works~\cite{Mao_2019_ICCV, Cui_2020_CVPR, Mao_2020_ECCV,Zang_IJCAI_2020} built spatial graphs over body-joints in each frame; however, such a single-scale modeling cannot easily capture a functional group of joints or high-order relations. For example, while walking, multiple joints on arms and legs collaborate together. Thus the modeling requires representations at multiple body scales, which are also adaptive to the input data.

In this work, we propose a~\emph{multiscale spatio-temporal graph} to comprehensively model human motions; see a visualization example in Fig.~\ref{fig:Jewel3}. This graph structure has three key features: decomposability, multiscale representation, and data adaptation. First, to reduce complexity and capture clear information, we decompose a spatio-temporal as one spatial graph and one temporal graph.  Second, we learn a~\emph{multiscale spatial graph} and a~\emph{multiscale temporal graph}: 1) a multiscale spatial graph consists of sub-graphs that represent the same pose at various spatial scales, where each vertex models one functional group of body-joints and each edge models vertices' relations; 2) a multiscale temporal graph consists of several graphs that represent the same sequences at various temporal scales, where each vertex models a time period and each edge models the temporal consistency. Third, the proposed multiscale spatial graph is trainable and adjusted during training. It is also dynamic in various network layers to reflect flexible relations.

\begin{figure*}[t]
    \centering
    \includegraphics[width=1.9\columnwidth]{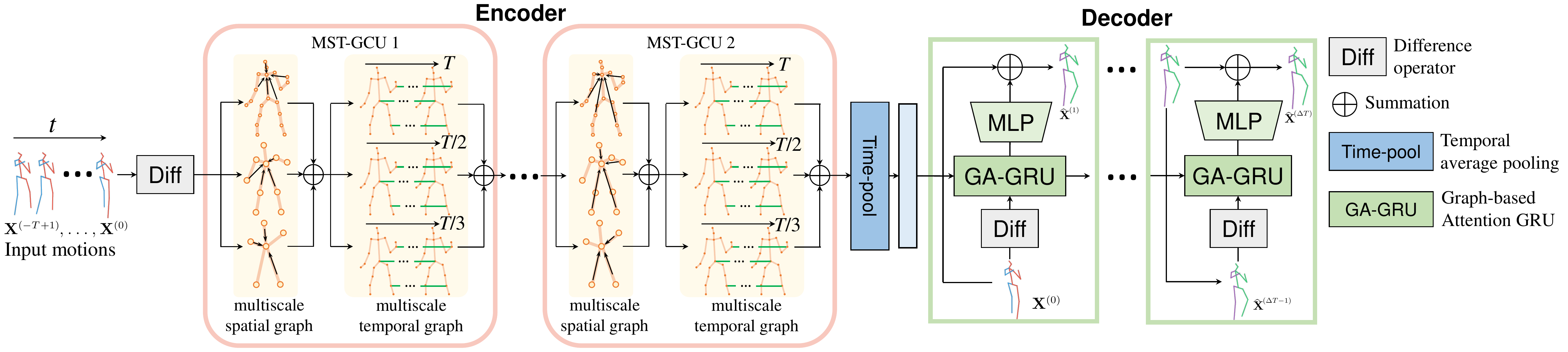}
    \caption{The architecture of MST-GNN, which uses an encoder-decoder framework for motion prediction. In the encoder, cascaded multiscale spatio-temporal graph computational units (MST-GCU) alternately leverage muliscale spatial and temporal graphs to extract spatio-temporal features. In the decoder, we propose a graph-based attention gated recurrent unit (GA-GRU) to sequentially predict poses in the future.}
    \vspace{-5mm}
    \label{fig:pipeline}
\end{figure*}

Based on the multiscale spatio-temporal graph, we develop a novel~\emph{multiscale spatio-temporal graph neural networks} (MST-GNN) to forecast the future 3D poses without action-categories. The proposed MST-GNN follows an encoder-decoder framework, where the encoder extracts high-level historical features and the decoder generates the future poses. In the encoder, the core module is a novel~\emph{multiscale spatio-temporal graph computational units} (MST-GCU), which extracts motion features given the multiscale spatio-temporal graph. Each MST-GCU includes three steps: multiscale graph construction, single-scale graph convolution, and cross-scale feature fusion. Each step further corresponds to the operations in both spatial and temporal domains. To achieve multiscale graph construction and cross-scale feature fusion, we propose novel graph pooling and unpooling operators to adaptively handle the irregular graph structure and maps between the body-joints and the body-components across arbitrarily scales. In the decoder, we propose a novel~\emph{graph-based attention gated recurrent unit} (GA-GRU) to sequentially produce predictions. The GA-GRU utilizes attentions based on trainable graphs to enhance the most important body-joint features. To learn richer motion dynamics, we introduce difference operators to extract multiple orders of motion differences as the proxies of positions, velocities, and accelerations. The architecture of the proposed MST-GNN is illustrated in Fig.~\ref{fig:pipeline}. To train the MST-GNN, we introduce a loss function that combines three terms: a prediction loss based on the $\ell_1$ distance between the predictions and ground-truths, a gram-matrix loss that minimizes the distance between the motion covariances of the predictions and ground-truths, and an entropy loss that promotes a clear clustering during graph pooling.

To verify our MST-GNN for human motion prediction, extensive experiments are conducted on three large-scale datasets: Human 3.6M~\cite{6682899} CMU Mocap\footnote{http://mocap.cs.cmu.edu/} and 3DPW~\cite{Marcard_2018_ECCV}. The experimental results show that our MST-GNN outperforms most state-of-the-art methods for short and long-term motion prediction.
The main contributions are as follows: 
\begin{itemize}
    \item We propose multiscale spatio-temporal graph neural networks (MST-GNN) to extract motion features at various scales and achieve effective motion prediction;
    \item We propose two key modules: a multiscale spatio-temporal graph computational unit (MST-GCU), which leverages trainable multiscale spatio-temporal graphs to extract motion features, and a graph-based attention GRU (GA-GRU) to enhance the pose generation; 
    \item We conduct extensive experiments to show that our model outperforms most state-of-the-art works by $5.33\%$ and {$3.67\%$} of mean angle errors for short-term and long-term prediction on Human 3.6M, and by {$11.84\%$} and {$4.71\%$} of mean angle errors for short-term and long-term prediction on CMU Mocap, respectively.
\end{itemize}

{
Compared to our previous work, DMGNN~\cite{Li_2020_CVPR}, the MST-GNN has three major technical improvements:}
\begin{itemize}
    \item 
    This work proposes a trainable multiscale spatio-temporal graph to model the motion data; while~\cite{Li_2020_CVPR} only considers multiscale spatial graphs, not fully modeling temporal information. Furthermore,  to obtain spatial graphs at various scales, we propose trainable spatial graph pooling and unpooling to infer cross-scale mappings; while~\cite{Li_2020_CVPR} uses predefined initial graph structures. 

    \item 
    This work proposes a graph-based attention GRU to improve the decoder in DMGNN~\cite{Li_2020_CVPR}. Based on the attention, we enhance the joint features that carry the most important dynamics for motion prediction.
    
    \item This work considers a series of losses to make the prediction more precise, including an $\ell_1$-based loss, a gram-matrix loss and an entropy loss; while ~\cite{Li_2020_CVPR} only considers the $\ell_1$-based prediction loss.
\end{itemize}

The rest of the paper is organized as follows: 
In Section~\ref{sec:Relatedwork}, we review some works related to human motion prediction and graph representation learning. 
In Section~\ref{sec:Problem}, we formulate the problem of human motion prediction and introduce some mathematical foundations of our model. 
In Section~\ref{sec:Keycomp}, we propose a key module in our network: multiscale spatio-temporal graph computational unit. 
In Section~\ref{sec:DMGNN}, we propose the dynamic multiscale graph neural network and the objective function.
Finally, the experiments validating the advantages of our model and the conclusion of the paper are provided in Section~\ref{sec:Experiments} and Section~\ref{sec:Conclusion}.

\section{Related Work}
\label{sec:Relatedwork}
\subsection{Human Motion Prediction}
  3D skeleton-based human motion prediction is a critical task that has been explored for a long time due its wide applications in real world. In the early period, many attempts developed traditional algorithms based on state models, such as hidden Markov models~\cite{Lehrmann_2014_CVPR}, Gaussian process models~\cite{NIPS2006_3078}, restricted Boltzmann machines~\cite{icml2009_129} and linear dynamics systems~\cite{NIPS2000_ca460332}, have proved their abilities. In recent years, deep learning techniques come to the forefront, which learn deep and flexible patterns to address this sequence-to-sequence motion prediction problem. As effective learning methods on sequential data, some recurrent-neural-network-based are developed. Encoder-Recurrent-Decoder (ERD)~\cite{Fragkiadaki_2015_ICCV} bridges a pair of nonlinear encoder and decoder with a recurrent feature learner. Structural-RNN~\cite{Jain_2016_CVPR} builds fully-connected graphs between body-parts and apply recurrent networks to propagate body-part-wise information. Pose-VAE~\cite{Walker_2017_ICCV} constructs a VAE with LSTM-based encoder and decoder.  Res-sup~\cite{Martinez_2017_CVPR} with RNN-based encoder-decoder constrains to model the pose displacement frame-by-frame for stable prediction. AGED~\cite{Gui_2018_ECCV} alters the Euclidean loss in \cite{Martinez_2017_CVPR} to a new loss considering the geometric structure and distribution. HMR~\cite{Liu_2019_CVPR} incorporates the Lie algebra in a hierarchical recurrent network to constrain the contextual features. DMGNN~\cite{Li_2020_CVPR} builds dynamic multiscale spatial graphs to capture the patterns of joint groups.
  
  Another type of motion prediction frameworks are feed-forward networks, which directly use feed-forward operations on entire motion sequences without remembering frame-wise history states. One important work is CSM~\cite{Li_2018_CVPR}, which builds both short-term and long-term encoder with spatio-temporal convolutions and uses convolution-based decoder to generate poses. TrajGCN~\cite{Mao_2019_ICCV} transforms the motion dynamics to the frequency domain by discrete cosine transform (DCT) and applies deep graph convolution on the trajectory representation. LDR~\cite{Cui_2020_CVPR} proposes a deep network with hierarchical feature extractor based on the spatial and temporal graphs. HisRep~\cite{Mao_2020_ECCV} builds a self-attention mechanism along time to emphasize periodic motion patterns for precise prediction. LPJP ~\cite{Cai_ECCV_2020} designs progressive information propagation strategies in a transformer-based networks. Compared to the previous works, our MST-GNN presents a new representation for human motion by building purely data-driven multiscale spatio-temporal graphs. Our model leverages such a graph to learn the comprehensive and informative semantics for more effective motion prediction.

  \subsection{Graph Representation Learning}
  Graphs effectively represent a large amount of data with non-grid structures and explicitly depict the correlations between vertices~\cite{AAAI1817135, 8779725}, which could be used to address various problems, such as social networks analysis~\cite{tabassum2018social}, medicine design or discovery~\cite{jin2018junction,you2018graph}, traffic forecasting~\cite{cao2021spectral,Hu_2020_CVPR,guo2019attention}, and human behavior modeling~\cite{Qi_2018_ECCV,Wang_2020_CVPR}. To capture the patterns based on both graph structures and vertex features, some methods adopted deep embedding based on random walk approximation~\cite{Perozzi_KDD2014,Grover_KDD2016} or small range proximities~\cite{Tang_WWW2015}. Recently, many works studied the graph neural networks (GNNs), generalizing the deep network to the graph domain. With hierarchical architectures and end-to-end training fashion, GNNs mainly fall into two perspectives: a spectral perspective and a vertex perspective. From a spectral perspective, graphs are converted into its spectrum signal processing~\cite{8745502}. For example, Spectral CNN~\cite{Bruna2014ICLR} used the eigen-decomposition of graph Laplacian. ChebyNet~\cite{NIPS2016_6081} used Chebyshev polynomial to approximate the convolution filters. Graph Convolution Network (GCN)~\cite{kipf_iclr2017} reduced the ChebyNet and combined the spectral analysis and spatial operation. From a vertex perspective, feature aggregation on graphs is directly designed, resembling the convolution on images~\cite{He_2016_CVPR}. Furthermoren, some works modified the GNNs, including randomly sampling neighboring nodes, learning edge attentions or building recurrent neural networks~\cite{NIPS2017_6703,pmlr-v70-Niepert16,velickovic2018graph,Li2016ICLR,ICML_2016_Dai}.
  
  Given the algorithms of graph representation learning, many related and practical tasks are explored, such as skeleton-based action recognition~\cite{8842613,AAAI1817135, Li_cvpr_2019,Shi_2019_CVPR,Hu_ICME_2019}, skeleton-based motion prediction~\cite{Mao_2019_ICCV,Cui_2020_CVPR,Cai_ECCV_2020,Li_2020_CVPR,Mao_2020_ECCV}, visual and scene reasoning~\cite{9062552,9067002,9102429,Fan_2019_ICCV,Qi_2018_ECCV,Wang_2020_CVPR,Lu_ECCV_2020} as well as multi-agent modeling~\cite{kipf2018neural,yu2020spatio,huang2019stgat,kosaraju2019social,li2020evolvegraph} . Compared to previous works, the proposed MST-GNN considers a multiscale spatio-temporal graph and its related operations, including graph pooling, graph convolution and graph unpooling to effectively extract motion features and improve motion prediction.

\section{Problem Formulation and Foundations}
\label{sec:Problem}
3D skeleton-based human motion prediction aims to generate a sequence of poses in the future guided by the observations.  Mathematically, let 
$\mathbf{X}^{(t)}\in\mathbb{R}^{M \times 3}$ 
be a pose matrix that records the 3D coordinates of $M$ body joints at time $t$,  
${\mathbb{X}} = [\mathbf{X}^{(1)},\dots,\mathbf{X}^{(T)}] \in \mathbb{R}^{T \times M \times 3}$ 
be a three-mode tensor that concatenates the pose matrices in a sequence of $T$ timestamps, where 
${\mathbb{X}}^{[t,s,c]}$ 
is the $c$th coordinate value of the $s$th body-joint at timestamp $t$.
Therefore, we could let 
${\mathbb{X}^{-}} = [\mathbf{X}^{(-T+1)},\dots,\mathbf{X}^{(0)}] {\in \mathbb{R}^{ T \times M \times 3}}$
be a tensor that represents $T$ historical poses, 
${\mathbb{X}^{+}} = [\mathbf{X}^{(1)},\dots,\mathbf{X}^{(\Delta T)}] {\in \mathbb{R}^{\Delta T \times M \times  3}}$ 
be a tensor that represents $\Delta T$ future poses. In motion prediction, we aim to propose a trainable predictor $\mathcal{F}_{\rm pred}(\cdot)$, which generates a sequence of the predicted poses $\widehat{\mathbb{X}}^{+}=\mathcal{F}_{\rm pred}(\mathbb{X}^{-})$ to approximate the ground-truth $\mathbb{X}^{+}$.

\vspace{-2mm}
\subsection{Necessity of trainable multiscale spatio-temporal graph}
\label{sec:necessity_multiscale}
Here we consider a~\emph{multiscale spatio-temporal graph} to model the spatio-temporal dependencies among of a skeleton-based human motion. 
The intuitions are three-folds. First, human bodies are regularized by some spatial constraints during moving; secondly, poses across consecutive timestamps are inertial and correlated; thirdly, many motions require the participation of a functional group of joints or consist of several temporal segments. 
These three observations reflect that a multiscale spatio-temporal graph could be introduced to learn the dynamics of human behaviors.

To define the multiscale graph, we first introduce a spatio-temporal graph at the original joint scale. Mathematically, $G_0(\mathcal{V}_0, \mathcal{E}_0, {\bf A}_0)$ is defined as a spatio-temporal graph that models the inter-joint relations in a motion, where $\mathcal{V}_0$ is the vertex set with $|\mathcal{V}_0|=TM$ joints; $\mathcal{E}_0$ is the edge set containing the spatio-temporal relations; and ${\bf A}_0 \in \mathbb{R}^{(TM) \times (TM)}$ is the graph adjacency matrix. We can rearrange $\mathbb{X}$ and combine the first two dimensions to form a pose matrix supported on this spatio-temporal graph, $\mathcal{X} \in \mathbb{R}^{(TM) \times 3}$.

Based on the spatio-temporal graph at the joint scale, a multiscale spatio-temporal graph consists of $R+1$ nested graphs. Besides the original $G_0(\mathcal{V}_0, \mathcal{E}_0, {\bf A}_0)$, we abstract another $R$ graphs as $G_1(\mathcal{V}_1, \mathcal{E}_1, {\bf A}_1) \dots G_R(\mathcal{V}_R, \mathcal{E}_R, {\bf A}_R)$, which contain $T_1 M_1, \dots, T_R  M_R$ vertices, respectively ($M_{r+1}<M_r$ and $T_{r+1}< T_r$, for $\forall~1 \leq r < R-1$). Each vertex in $\mathcal{V}_r$ means a group of body joints in the $r$th scale. The details about multiscale graph construction will be elaborated in Section~\ref{sec:Keycomp}.

There are two challenges to construct a multiscale spatio-temporal graph structure: 1) fixed spatio-temporal graph cannot adapt to numerous human motions, since there are implicit and action-related constraints on bodies during moving, which are hard to be manually determined. 2) the spatio-temporal connections might form a large-size graph, being hard to store and process in the real-time prediction.

For the first challenge, we propose the trainable multiscale spatio-temporal graphs to capture highly flexible correlations in both spatial and temporal domains; that is, the graph adjacency matrix at each single scale is adaptively adjusted during training to maximally fit the implicit relations in motions. Additionally, the graph construction across scales are also trainable. Furthermore, in each network layer, the associated multiscale spatial-temporal graph structure is trained individually, thus it has more flexibility to capture different relations on hierarchical features. 

\begin{figure}[t]
    \centering
    \includegraphics[width=0.8\columnwidth]{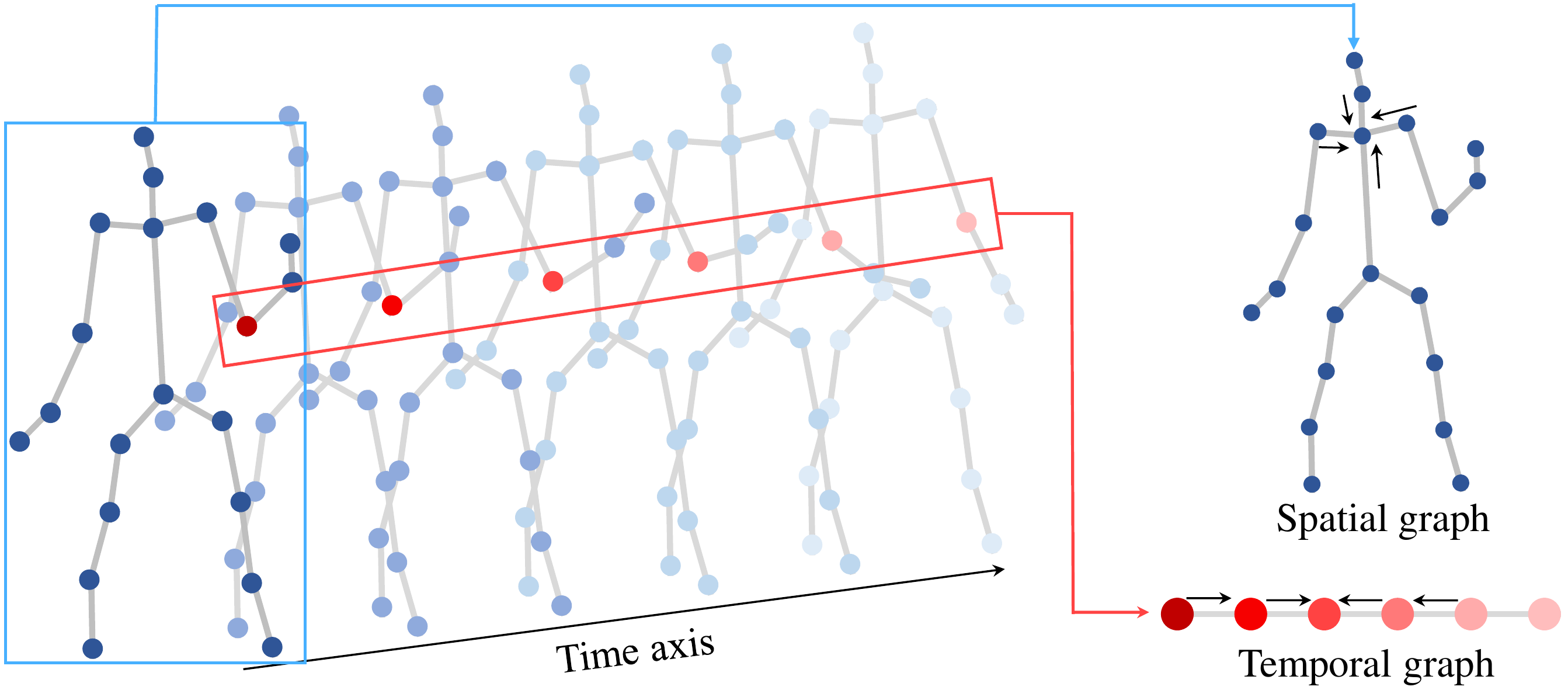}
    \small
    \caption{A sketch of the spatio-temporal graph to represent the motion. We consider a motion sequence separately as a spatial graph and a temporal graph, on which the vertex information is propagated. The spatial graph depicts the spatial relations on a human body at a frame; and the temporal graph depicts the dependencies of poses along time.}
    \vspace{-5mm}
    \label{fig:SpatioTemporalGraph}
\end{figure}

To address the second challenge, we consider a \emph{decomposable assumption} for the spatio-temporal graph; that is, at any scale $r$, we decompose the hybrid spatio-temporal graph into a spatial graph and a temporal graph. We consider the adjacency matrix ${\bf A}_r$ as the Cartesian product of a spatial graph and a temporal graph: 
  $  {\bf A}_r = {\bf S}_{r} \otimes {\bf T}_{r},$
where ${\bf S}_{r}\in\mathbb{R}^{M_r \times M_r}$ and ${\bf T}_{r} \in \mathbb{R}^{T_r \times T_r}$ are the adjacency matrices of spatial and temporal graphs, respectively, and $\otimes$ is the Cartesian product. ${\bf S}_{r}$ reflects the spatial relations among joints at a frame and ${\bf T}_{r}$ reflects the temporal dependencies along time at a joint. The trainable ${\bf S}_{r}$ and ${\bf T}_{r}$ are initialized according to either a skeleton structure or temporal priors, and then, the graph structures are adjusted adaptively; see a visualization in Fig.~\ref{fig:SpatioTemporalGraph}. {Such a decomposition not only reduces the computational and storage costs, but it also allows us to design distinct operations on either the spatial or temporal graphs, which model different types of relations to learn corresponding patterns.}

\vspace{-3mm}
\subsection{Basics of spatial/temporal graph convolution}
To extract features from a spatio-temporal graph at any scale, we propose a generic spatio-temporal graph convolution. Considering a spatio-temporal graph ${\bf A} = {\bf S} \otimes {\bf T} \in \mathbb{R}^{(TM) \times (TM)}$ at a single scale, and we let $\mathbb{X} \in \mathbb{R}^{T \times M \times D}$ be a motion tensor and $\mathcal{X} \in \mathbb{R}^{(TM) \times D}$ be the spatio-temporal matrix. Based on the decomposable assumption, we define the spatio-temporal graph convolution in two equivalent forms,
\begin{subequations}
\setlength{\abovedisplayskip}{4pt}
\setlength{\belowdisplayskip}{4pt}
\begin{eqnarray}
\label{eq:joint_convolution}
\mathcal{X}' & = & {\bf F} *_{{\bf A}} \mathcal{X} 
= {\bf F} *_{{\bf S} \otimes {\bf T}} \mathcal{X},
\\
\label{eq:separate_convolution}
\mathbb{X}' & = &  \mathbb{V} *_{{\bf T}}  \left( \mathbb{U} *_{{\bf S}} \mathbb{X} \right),
\end{eqnarray}
\end{subequations}
where $*_{{\bf S} \otimes {\bf T}}$ denotes the generic spatio-temporal graph convolution for $\mathcal{X}$, $*_{{\bf S}}$ and $*_{{\bf T}}$ denote the decomposed spatial graph convolution and temporal graph convolution, respectively, and ${\bf F},  \mathbb{U}, \mathbb{V}$ denote the graph filters.~\eqref{eq:joint_convolution} means that the filter ${\bf F}$ convolves the input $\mathcal{X}$ on the spatio-temporal graph ${\bf A}$; and~\eqref{eq:separate_convolution} means that a spatio-temporal graph convolution can be decomposed into a spatial graph convolution followed by a temporal graph convolution.

Based on~\eqref{eq:separate_convolution}, we respectively formulate the spatial and temporal graph convolution in detail. First, the spatial graph convolution handles the data at each frame independently. For the slice of the $t$th timestamp in $\mathbb{X}$, it works as 
\begin{equation}
\label{eq:spatial_convolution}
\setlength{\abovedisplayskip}{4pt}
\setlength{\belowdisplayskip}{4pt}
\left( \mathbb{U} *_{{\bf S}} \mathbb{X} \right)^{[t,:,:]} 
= \sum_{\ell=0}^{L} {\bf S}^{\ell} \mathbb{X}^{[t,:,:]}  {\bf U}_{\ell} \ \in \ \mathbb{R}^{M \times D'},
\end{equation}
where $\mathbb{U}  \in \mathbb{R}^{L \times D \times D'}$ with the $\ell$th slice ${\bf U}_{\ell} \in \mathbb{R}^{D \times D'}$ is a matrix of trainable filter coefficients corresponding to the $\ell$th order, which cover the corresponding reception fields on a graph. Accordingly, the temporal graph convolution handles the data at each spatial vertex. For the slice of the $s$th spatial vertex in $\mathbb{X}$, it works as 
\begin{equation}
\label{eq:temporal_convolution}
\setlength{\abovedisplayskip}{4pt}
\setlength{\belowdisplayskip}{4pt}
\left( \mathbb{V} *_{{\bf T}} \mathbb{X} \right)^{[:,s,:]} 
= \sum_{\ell=-L}^{L} {\bf T}^{\ell} \mathbb{X}^{[:,s,:]}   {\bf V}_{\ell} \ \in \ \mathbb{R}^{T \times D'},
\end{equation}
where $\mathbb{V} \in \mathbb{R}^{2L \times D \times D'}$ with the $\ell$th slice ${\bf V}_{\ell}$ is a matrix of trainable graph filter weights that corresponds to the $\ell$th order.

The spatial and temporal graph convolution can be alternately used and effectively extract rich features, therefore, the spatial and temporal graph convolution does not need fixed order of calculation in practice, such as applying \eqref{eq:spatial_convolution} and then \eqref{eq:temporal_convolution}, because the iterative operations could embed highly hybrid spatio-temporal information. 
In our method, we first apply \eqref{eq:spatial_convolution} for convenience, while we test two orders in our experiments to verify the similarity; see results in Appendix.

\section{Multiscale Spatio-Temporal Graph Computational Unit}
\label{sec:Keycomp}
In this section, we propose a core component in our network, called~\emph{multiscale spatio-temporal graph computational unit} (MST-GCU). The key of MST-GCU is 
to leverage a trainable multiscale spatio-temporal graph to represent the input data and then design the operations to extract features in every single scale and across scales. Following by Section~\ref{sec:necessity_multiscale}, our multiscale spatio-temporal graph should be decomposable, leading to spatial and temporal graphs at each scale to reduce the computational cost and clearly represent the two types of relations. Accordingly, we design single-scale graph convolution and cross-scale fusion operations for spatial and temporal graphs.

\begin{figure}[t]
    \centering
    \includegraphics[width=0.95\columnwidth]{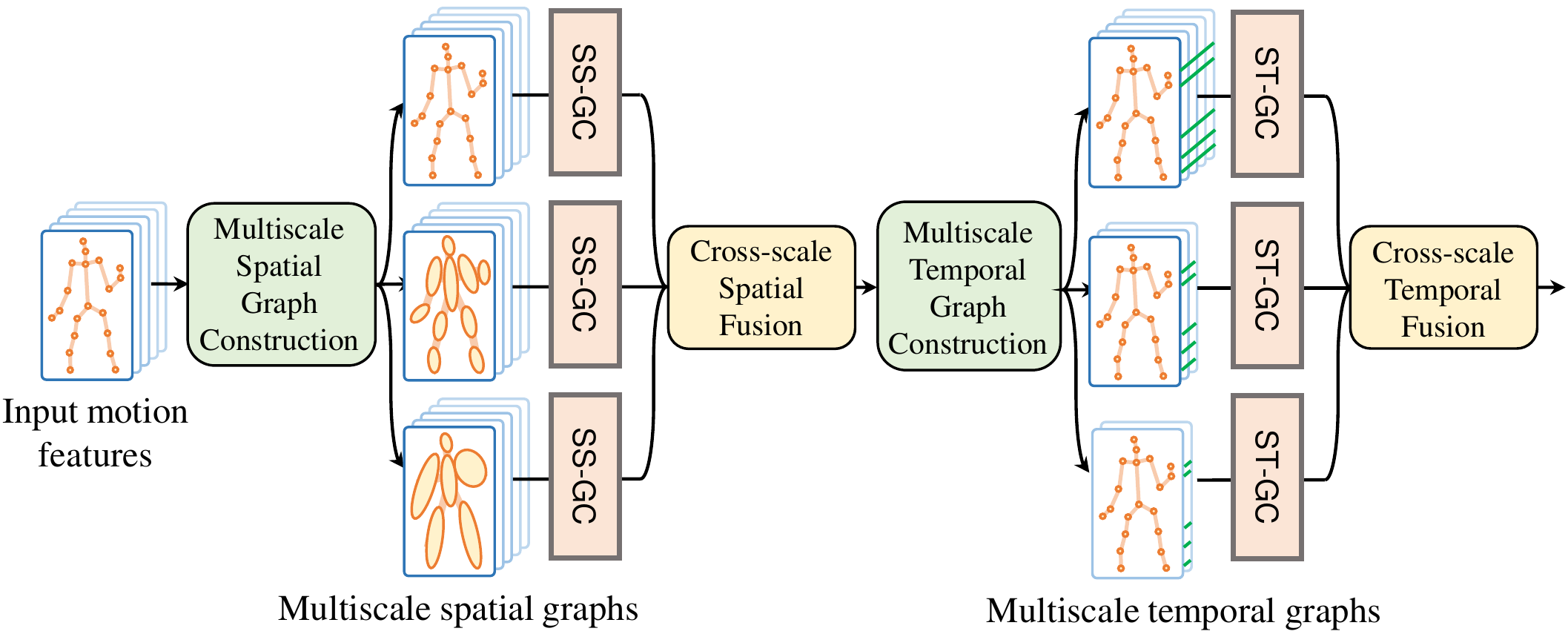}
    \vspace{-3mm}
    \small
    \caption{An MST-GCU learns comprehensive motion features from multiscale spatial graph and multiscale temporal graph.}
    \label{fig:MGCU}
    \vspace{-5mm}
\end{figure}

Overall, MST-GCU includes three steps: multiscale graph construction, graph convolution in each scale and cross-scale feature fusion; see Fig.~\ref{fig:MGCU}. We next introduce the operations.

\vspace{-3mm}
\subsection{Multiscale Spatial/Temporal Graph Construction}

\begin{figure}[t]
    \centering
    \includegraphics[width=0.85\columnwidth]{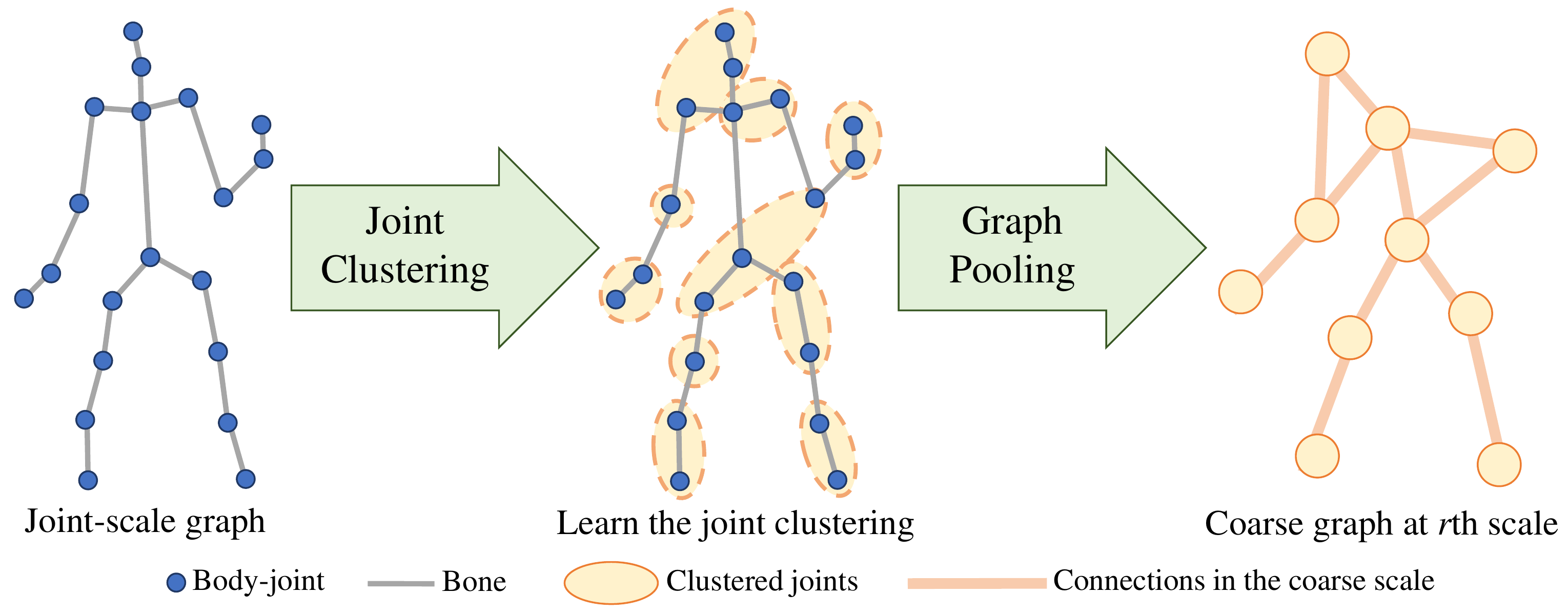}
    \vspace{-3mm}
    \small
    \caption{The multiscale spatial graphs generated on Human 3.6M. We note that, the spatial graph at each scale is purely trainable, while we only show the initial skeleton-based structure at the joint-scale for visualization.}
    \label{fig:multiscaleDSgraph}
    \vspace{-5mm}
\end{figure}

\subsubsection{Multiscale spatial graph construction}
We now present how to construct multiscale spatial graphs.
We first build the spatial graph in the original scale, ${\bf S}_0$, which is initialized according to the skeletal connections: the vertices are joints and edges are bones. Then, the elements of ${\bf S}_0$ are adjusted adaptively during model training to reflect implicit relations.

To adaptively coarsen spatial graphs at various scales, we propose a trainable~\emph{spatial-graph pooling operator}, which maps the original spatial graph to an arbitrary scale by grouping correlated body joints; see Fig.~\ref{fig:multiscaleDSgraph}. The spatial-graph pooling operator is parameterized and trained in an end-to-end manner. Let $\mathbb{X} \in\mathbb{R}^{T \times M \times d}$ be the spatio-temporal data in the original scale, ${\bf S}_0$ and ${\bf T}_0$ be the spatial and temporal graph adjacency matrices. In the $r$th spatial scale, the spatial-graph pooling operator ${\Psi}_{0 \to r} \in [0,1]^{M \times M_r}$ is formulated as
\begin{equation}
\setlength{\abovedisplayskip}{4pt}
\setlength{\belowdisplayskip}{4pt}
    \Psi_{0 \to r} = 
    \sigma
    \bigg(
     {\bf S}_0
    \Big[
    {\rm ReLU} 
    \left ( 
    \mathbb{V} *_{{\bf T}_0} \mathbb{X}
    \right ) 
    \Big]_{13} {\bf W}_{0 \to r}
    \bigg),
\end{equation}
where $*_{{\bf T}_0}$ follows from~\eqref{eq:temporal_convolution}, $[\cdot]_{13}:\mathbb{R}^{T \times M \times d} \to \mathbb{R}^{M \times (dT)}$ reshape to merge the first and third tensor dimensions, ${\bf W}_{0 \rightarrow r} \in\mathbb{R}^{(dT) \times M_r}$ is the trainable weights;
and ${\sigma}(\cdot)$ is a softmax operation on each row to normalize the pose features and enhance the strongest responses. $({\Psi}_{0 \rightarrow r})_{i,j}$ indicates that the $i$th body joint in the original scale should assign to the $j$th groups in the $r$th spatial scale. The advantages of learning ${\Psi}_{0 \to r}$ include i) we leverage both spatial and temporal information to infer ${\Psi}_{0 \to r}$ across two spatial scales;  and ii) the learned ${\Psi}_{0 \to r}$ at various network layers are quite different, improving the flexibility of multiscale graph construction.

With ${\Psi}_{0 \to r}$, we convert the original spatial features and spatial graph adjacency matrix to the $r$th spatial scale:
\begin{subequations}
\setlength{\abovedisplayskip}{4pt}
\setlength{\belowdisplayskip}{4pt}
 \begin{eqnarray}
      \label{eq:x_ass_s}
      \mathbb{X}^{[t,:,:]}_r & = &  \Psi_{0 \rightarrow  r}^{\top}  \mathbb{X}^{[t,:,:]}, \\
      \label{eq:a_ass_s}
      {\bf S}_{r} & = & \Psi_{0 \rightarrow r }^{\top}{\bf S}_0 \Psi_{0 \rightarrow r },
\end{eqnarray}
\end{subequations}
where~\eqref{eq:x_ass_s} fuses the features of multiple body joints to form the feature of a body component in the $r$th scale at each individual timestamp; and correspondingly,~\eqref{eq:a_ass_s} coarsens the original spatial graph to obtain new connections in the $r$th scale.
In this way, the spatial-graph pooling operator effectively generates the coarse spatial features and spatial graphs. $\mathbf{S}_r$ in each MST-GCU is trained individually to show flexible relations, while it is fixed during test.

Notably, any scales of graphs are inferred from the original scale. The intuition is that the multiscale graphs meanwhile carry different perspectives of pose representation, thus we independently construct the multiscale graph from the original graph rather than consider a pyramid coarsening.

Some previous works of spatial graph pooling were also proposed. 
DMGNN~\cite{Li_2020_CVPR} employs human priors, where the expressiveness and applicability of a fixed skeleton structure are often limited. 
DGM~\cite{Li_2020_CVPR_DGM} constructs a node affinity matrix, which is computed based on a Gaussian kernel function to measure the distance between nodes across scales. gVAE~\cite{Dhamala_2019} uses a binary assignment matrix and clusters nodes to minimize the normalized cuts.  Compared to DMGNN, DGM and gVAE, our MST-GCU uses a feed-forward spatio-temporal graph convolution to directly learn the affinity matrix, which are trainable and efficient, as well as enable more flexible multiscale representation learning.

\subsubsection{Multiscale temporal graph construction}
Here we construct multiscale temporal graphs.
In the original scale, we initialize the temporal graph ${\bf T}_0$ by the standard sequential connections, which is a cyclic shift matrix,
\begin{equation}
\label{eq:cyclic}
\setlength{\abovedisplayskip}{4pt}
\setlength{\belowdisplayskip}{4pt}
    {\bf T}_0 \leftarrow 
    \left[
      \begin{matrix}
      0 & 0  & \cdots & 0 & 1      \\
      1 & 0  & \cdots & 0 & 0      \\
      \vdots & \vdots & \ddots & \vdots & \vdots\\
      0 & 0  & \cdots & 0 & 0      \\
      0 & 0  & \cdots & 1 & 0      \\
\end{matrix}
\right] \in \mathbb{R}^{T \times T}.
\end{equation}
${\bf T}_0$ reflects that each body-joint connects to the corresponding joint at the last timestamp; see a sketch at the bottom-right of Fig.~\ref{fig:SpatioTemporalGraph}, where we only show the elbow connections for visualization. To make ${\bf T}_0$ trainable, we parameterize all the nonzero elements in~\eqref{eq:cyclic} and fix all the zero elements. Therefore, we adaptively adjust the relationship strengths between consecutive frames.

To construct the temporal graph at various scales, we propose a~\emph{temporal-graph pooling operator}, which averages groups of several consecutive frames as new frames in a coarsened scale. The intuition is that the consecutive frames have similar poses and continuous evolution, thus we can consider them in a more abstract state. Mathematically, in the $r$th temporal scale, the temporal-graph pooling operator ${\Phi}_{0 \to r}\in[0,1]^{T \times T_r}$ is defined as
\begin{equation}
\setlength{\abovedisplayskip}{4pt}
\setlength{\belowdisplayskip}{4pt}
    \Phi_{0 \rightarrow r } \ = \ 
    \left [
    \begin{matrix}
    T_r/T & 0 & \cdots  & 0 \\
    T_r/T & 0 & \cdots  & 0 \\
    \vdots  & \vdots  &   \ddots  & \vdots\\
    0 & 0 & \cdots  & T_r/T \\
    0 & 0 & \cdots  & T_r/T
    \end{matrix}
    \right ],
\end{equation}
where each row has $T/T_r$ nonzero entries and sums up to one.

With ${\Phi}_{0 \to r}$, we convert the temporal features and temporal graph adjacency matrix to the $r$th scale; that is,
\begin{subequations}
\setlength{\abovedisplayskip}{4pt}
\setlength{\belowdisplayskip}{4pt}
 \begin{eqnarray}
      \label{eq:x_ass}
      \mathbb{X}^{[:,s,:]}_r & = &  \Phi_{0 \rightarrow  r}^{\top}  \mathbb{X}^{[:,s,:]}, \\
      \label{eq:a_ass}
      {\bf T}_{r} & = & \Phi_{0 \rightarrow r }^{\top}{\bf T}_0 \Phi_{0 \rightarrow r },
\end{eqnarray}
\end{subequations}
In this way, the temporal-graph pooling operator generates the coarsened temporal features and temporal graphs.

Overall, the spatial and temporal graphs at all scales are trainable. For the spatial graphs ${\bf S}_r$, both graph structures and edge weights are trainable; for the temporal graphs ${\bf T}_r$, graph structures are fixed to preserve the structure of the temporal sequence, but edge weights are trainable.

\vspace{-3mm}
\subsection{Single-Scale Spatial/Temporal Graph Convolution}
We next propose two graph convolution operators based on the spatial and temporal graphs. Since both are operated in every single scale; we thus name them~\emph{single-scale spatial graph convolution} (SS-GC) and~\emph{single-scale temporal graph convolution} (ST-GC), respectively.

\subsubsection{Single-scale spatial graph convolution}
To extract informative features along the spatial dimension in each scale, the single-scale spatial graph convolution (SS-GC) leverages the trainable spatial graph and aggregates neighboring information for each body component.  Let $\mathbb{X}_r \in \mathbb{R}^{T \times M_r \times D}$ be input spatio-temporal features at the $r$th spatial scale,  the output features is formulated as
\begin{equation}
\setlength{\abovedisplayskip}{4pt}
\setlength{\belowdisplayskip}{4pt}
\label{eq:SS-GC}
\mathbb{X}_{r, s} = 
{\rm ReLU}
\Big(  \mathbb{U} *_{{\bf S}_r} \mathbb{X}_r
\Big) \in\mathbb{R}^{T \times M_r \times D'},
\end{equation}
where $*_{{\bf S}_r}$ follows from~\eqref{eq:spatial_convolution}, indicating the convolution on ${\bf S}_r$, and the subscript $s$ in $\mathbb{X}_{r, s}$ specifies that the spatial features are updated. 
Note that i) features extracted at various scales could reflect information with different receptive fields; ii) both the graph filter coefficients in $\mathbb{U}$ and the underlying spatial graph ${\bf S}_r$ are trainable and vary in different MST-GCUs, reflecting dynamic spatial relations in various network layers.

\subsubsection{Single-scale temporal graph convolution}
To extract features along the temporal dimension in each scale, the single-scale temporal graph convolution (ST-GC) leverages the temporal graph to aggregate information from consecutive timestamps.  Given the input features at the $r$th temporal scale, $\mathbb{X}_r \in \mathbb{R}^{T_r \times M \times D}$,  the output features is
\begin{equation}
\setlength{\abovedisplayskip}{4pt}
\setlength{\belowdisplayskip}{4pt}
    \label{eq:ST-GC}
    \mathbb{X}_{r, t} = 
    {\rm ReLU}
    \Big(
    \mathbb{V}  *_{{\bf T}_r} \mathbb{X}_{r}
    \Big) \in\mathbb{R}^{T_r \times M \times D'},
\end{equation}
where $*_{{\bf T}_r}$ follows from~\eqref{eq:temporal_convolution} and the subscript $t$ indicates the temporal features are updated. The graph filter coefficients in $\mathbb{V}$ are trainable and vary in different MST-GCUs, reflecting dynamic temporal relations across network layers.

Previous GCLNC~\cite{Zhong_2019_CVPR} builds temporal graphs based on the temporal distances on a single scale. For a frame, the relations with the historical and future frames are fixed, symmetrical and undirected.
In our model, the temporal graph is asymmetrical over time. The intuition is that we consider the chronological effects from the history to forecast the future. Second, we consider multi-order relationship, where we not only build long-range edges, but we also use distinct and non-shared network parameters in graph convolution to model different orders of information. Finally, our temporal graph is built on multiple scales to capture the temporal information over the short- and long-term dynamics simultaneously.

\vspace{-3mm}
\subsection{Cross-Scale Spatial/Temporal Fusion}
After extracting spatial and temporal features in each scale, we next fuse those features across multiple scales to enhance the information flow and enable each layer of MST-GCU to carry the multiscale representation. Due to the different graph scales and structures, we cannot trivially use regular upsampling and summation to merge them. To address this issue, we propose the cross-scale spatial and temporal fusion to convert the coarsened spatial and temporal graph to the finest scale for information integration.

\subsubsection{Cross-scale spatial fusion.}
\begin{figure}[t]
    \centering
    \includegraphics[width=0.95\columnwidth]{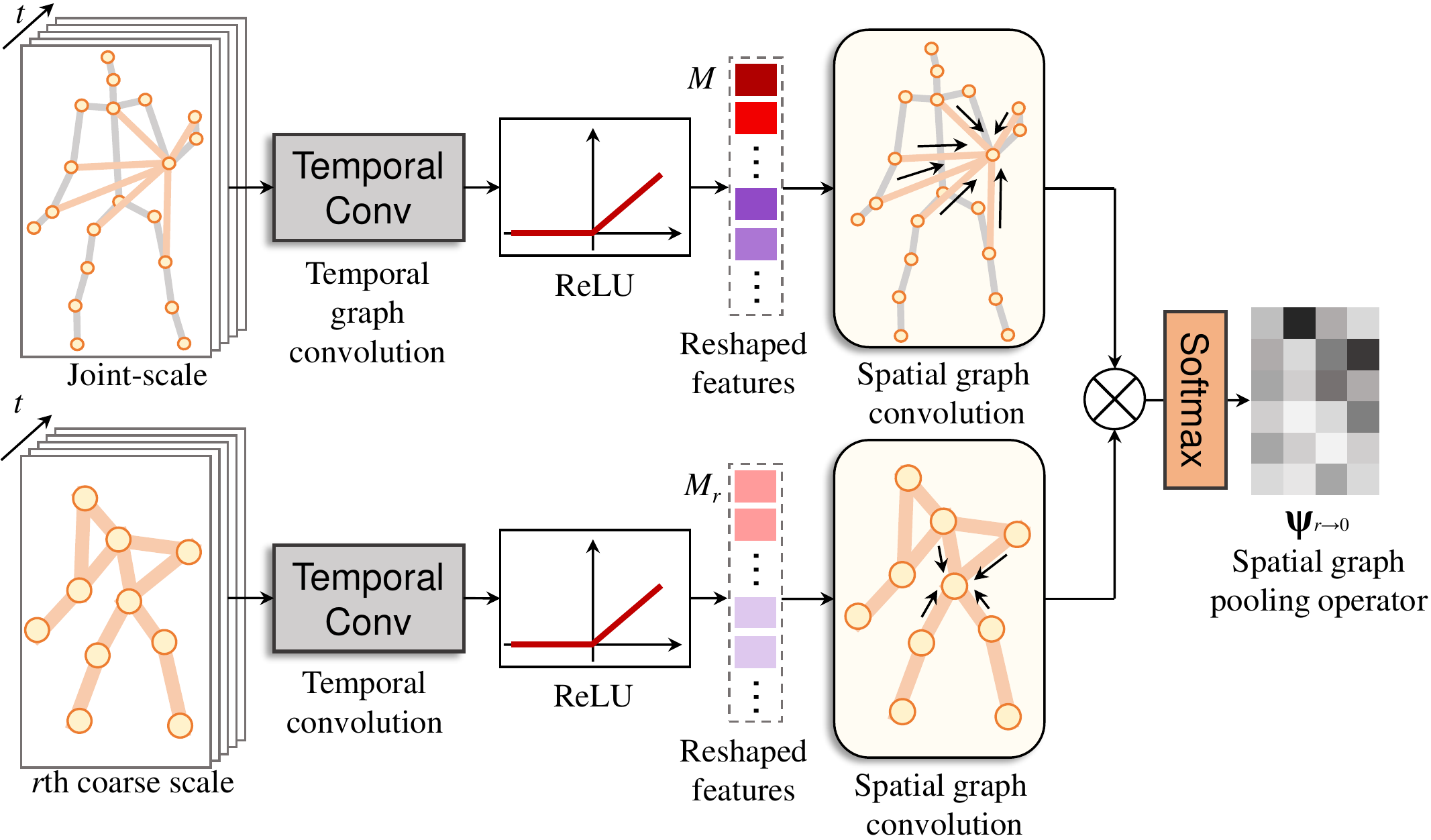}
    \vspace{-3mm}
    \small
    \caption{The inference of a cross-scale unpooling operator $\Psi_{r \to 0}$ between the joint-scale and the $r$th coarse scale for the cross-scale spatial fusion.}
    \vspace{-5mm}
    \label{fig:DC-FB}
\end{figure}

For cross-scale spatial features fusion, we propose a~\emph{cross-scale spatial fusion} unpooling operator to convert features in an arbitrary scale back to the original scale. Let 
${\mathbb{X}_{0}} \in \mathbb{R}^{T \times M \times D'}$ and
${\mathbb{X}_{r}} \in \mathbb{R}^{T \times M_r \times D'}$ be the features in the $0$th and $r$th spatial scales, respectively. We first embed the tensors in two spatial scales into two feature matrices,
\begin{subequations}
\setlength{\abovedisplayskip}{4pt}
\setlength{\belowdisplayskip}{4pt}
  \begin{align}
      {\bf Y}_0 & = 
      {\bf S}_0 
      \left [
      {\rm ReLU}
      \left (
      \mathbb{V}_0 *_{{\bf T}_0} \mathbb{X}_{0}
      \right )
      \right ]_{13}
      {\bf \Theta}_0,
      \\
      {\bf Y}_r & = 
      {\bf S}_r 
      \left [
      {\rm ReLU}
      \left (
      \mathbb{V}_r *_{{\bf T}_r} \mathbb{X}_{r}
      \right )
      \right ]_{13}
      {\bf \Theta}_r,
  \end{align}
\end{subequations}
where $\mathbb{V}_0, \mathbb{V}_r, {\bf \Theta}_0$ and ${\bf \Theta}_r$ are trainable weights,
$[\cdot]_{13}$ is the reshape operation to merge the $1$st and $3$rd tensor dimensions, and $*_{{\bf T}_r}$ follows from~\eqref{eq:temporal_convolution}. 
Then, we define the relations between any body joints in the original scale and any body components in the $r$th spatial scale as the relation matrix $\Psi_{r \to 0} \in[0,1]^{M \times M_r}$, whose elements is formulated as
\begin{equation}
\setlength{\abovedisplayskip}{4pt}
\setlength{\belowdisplayskip}{4pt}
    \label{eq:edge}
    \left( \Psi_{r \to 0} \right)_{i,j} 
    =
    \frac{
    \exp 
    \left (
    \left (
    {{\bf Y}_{0}}
    \right )_{i}^{\top} 
    \left (
    {{\bf Y}_r}
    \right )_{j}
    \right )
    }
    {
    \sum_{k=1}^{M}
    \exp
    \left (
    \left (
    {{\bf Y}_{0}}
    \right )_{i}^{\top}
    \left (
    {{\bf Y}_r}
    \right )_{k}
    \right )} 
    \in [0,1],
\end{equation}
where we use the inner product and a softmax to obtain the weight between the $i$th body component at the $r$th scale to the $j$th body joint. Fig.~\ref{fig:DC-FB} illustrates the inference of $\Psi_{r \to 0}$. 
Notably, $\Psi_{r \to 0}$ is efficiently inferred and adaptive to spatio-temporal features, which is flexible to capture distinct patterns from different input motions.

We finally fuse features from multiple spatial scales. At timestamp $t$, given the input $\mathbb{X}_r^{[t,:,:]} \in\mathbb{R}^{M_r\times d}$, the feature after cross-scale fusion is
\begin{equation}
\setlength{\abovedisplayskip}{4pt}
\setlength{\belowdisplayskip}{4pt}
    \label{eq:global_part}
    \mathbb{X}_0^{[t,:,:]}
    \gets
    \mathbb{X}_0^{[t,:,:]} + \sum_{r = 1}^R \Psi_{r \to 0} 
    \mathbb{X}_r^{[t,:,:]}
    \mathbf{W}_{r \to 0},
\end{equation}
where $\mathbf{W}_{s \to 0}\in\mathbb{R}^{d \times d}$ is trainable weights. Thus, each body joint in the original scale aggregates information from coarse spatial scales through $\Psi_{r \to 0}$.

\subsubsection{Cross-scale temporal fusion.}
According to the regular lattice structure of the multiscale temporal graph, we can use a convenient and effective method to fuse the temporal features across temporal scales. By copying consecutive frames, we fill in the multiscale temporal representations with equal intervals according to the sampling density when generating the multiscale temporal graphs. For example, when the coarse temporal graph shrinks the original temporal graph to one-half along time, for each preserved frame, we insert a duplicate of this frame before or after itself to enlarge the temporal scales and add the filled sequence to the original scale for fusion. To evaluate the effectiveness and importance of multiscale information, we conduct extensive experiments on model performance.

The intuition includes 1) the temporal graph and features are based on a regular grid structure, where we can convert temporal graphs from one coarse scale to a finer one straightforwardly, and we do not need extra parameters to increase the model complexity; 2) our fusion method effectively preserves the sequential order of human motion information.

\vspace{-3mm}
\subsection{Ensemble}
In previous subsections, we have introduced each building block of MST-GCU. Here we assemble all the building blocks to build up our MST-GCU module. As illustrated in Fig.~\ref{fig:MGCU}, MST-GCU first extracts spatial features by sequentially performing multiscale spatial graph construction, single-scale spatial graph convolution, and cross-scale spatial fusion; with those spatial features as input, MST-GCU further extracts temporal features by sequentially performing multiscale temporal graph construction, single-scale temporal graph convolution, and cross-scale temporal fusion. The design rationale includes: 
1) By decomposing the spatio-temporal graphs into separated spatial and temporal domains, we reduce the size of the giant spatio-temporal graph from $M^2 \times T^2$ to $M^2+T^2$. In this way, our method alleviates the problem that the giant spatio-temporal graph causes large storage and computational costs, improving the efficiency of the motion prediction for real-time application.
2) Spatial relations and temporal relations have naturally different physical meanings, which could constrain our MST-GCU to focus on either spatial or temporal information for clear feature extraction; on the contrary, confusing the spatial and temporal relationships in the same space without being aware of their differences might make difficult to train the model to capture accurate spatio-temporal feature.

Many previous works also leverage spatio-temporal graphs or multiscale/pyramid/hierarchical architectures to capture motion dynamics for prediction~\cite{Li_2018_CVPR,Gui_2018_ECCV, AAAI1817135}. Compared to them, the main novelty of the proposed network is that the trainable multiscale spatio-temporal graph and information. Concretely, compared to~\cite{Shi_2019_CVPR,Li_cvpr_2019}, their graph structure is trainable, yet represents on a single scale of a human body; while our graph represents on multiple scales, providing comprehensive information from different perspectives for motion prediction. As for~\cite{SymGNN,Si_2019_CVPR}, they leverage predesigned and fixed multiscale graphs; while we construct a trainable multiscale graph. To learn the multiscale information, we propose a series of operations in MST-GCU: multiscale graph construction, multiscale feature learning and multiscale feature fusion, which form a complete and effective scheme. The proposed key module, MST-GCU, could be regarded as an individual system that is potentially transferred to many other spatio-temporal representation learning problems.

\section{Network Architecture}
\label{sec:DMGNN}
Here we present the entire architecture of our MST-GNN, which contains a multiscale graph-based encoder and a recurrent graph-based decoder for motion prediction.

\vspace{-3mm}
\subsection{Encoder}
To capture the high-level semantic features from the observed sequence of human poses, the encoder targets to provide the decoder with informative motion states for prediction; see the left part in Fig.~\ref{fig:pipeline}. Let the input data be $\mathbb{X} \in \mathbb{R}^{T \times M \times 3}$, the encoder $\mathcal{E}(\cdot)$ produces the motion state representation: ${\bf H} \ = \ \mathcal{E}(\mathbb{X})\in\mathbb{R}^{M \times D_{\rm h}},$
where each body joints carries the individual dynamics for motion prediction.

In the encoder, we first utilize a graph convolution on the input motion data to extract some initial features for multiscale graph generation and representation learning. Based on the low-level spatial features computed by the first layer of graph convolution, we apply $L$ layers of multiscale spatio-temporal graph computational units (MST-GCU) to learn multiscale motion features. {Given the cascade of $L$ MST-GCUs, the encoder performs iteratively information propagation on both spatial and temporal domains to extract comprehensive features. After  each  multiscale  spatial/temporal  graph  convolution,  we  aggregate  the  multiscale  information  via  the  cross-scale  spatial/temporal  fusion.  We  note  that, the cross-scale  spatial/temporal  fusion is  not  the  end  of  information  propagation;  on  the  contrary,  the  cross-scale  fusion works on combining the multiscale information to enhance information flow; that is, the multiscale features are sufficiently propagated across scales and provide the comprehensive information for the downstream operation}

In each MST-GCU, we employ $3$ spatial scales of human poses and $3$ temporal scales of video sequences. To extract the multiscale features, the parallel single-scale spatial graph convolution (SS-GC) and single-scale temporal graph convolution (ST-GC) associated with the three scales are trained individually. For each MST-GCU, we additionally introduce a residual connection from its input to its output to improve the gradient backpropagation and accelerate training; if the dimensionalities of input and output are different, we adopt a 1D convolution on the input spatio-temporal feature to update the feature dimensionality. Finally, we use global average pooling on the temporal dimension to obtain the motion state matrix ${\bf H}\in\mathbb{R}^{M \times D_{\bf h}}$.

To make use of the motion dynamics, we not only utilize the spatial position of the human poses as the input data, but also consider the velocities and accelerations of joints, which carry crucial information of the movement process.  The velocity and acceleration are obtained through a difference operator, which could compute any high-order difference of the input sequence.
At time $t$, the $0$-order pose difference is $\Delta^{0}\mathbf{X}^{(t)}  = \mathbf{X}^{(t)}$, and the $\beta$-order pose difference ($\beta > 0$) is 
$
    \Delta^{\beta}\mathbf{X}^{(t)} = \Delta^{\beta-1}\mathbf{X}^{(t)}-\Delta^{\beta-1}\mathbf{X}^{(t-1)}.
$
We use zero paddings after computing the differences to handle boundaries. 
Overall, the $\beta$-order difference of the motion is
\begin{equation}
\setlength{\abovedisplayskip}{4pt}
\setlength{\belowdisplayskip}{4pt}
{\rm diff}_{\beta}(\mathbf{X}^{(t)})  =   \begin{bmatrix} \Delta^{0}\mathbf{X}^{(t)} & \cdots & \Delta^{\beta}\mathbf{X}^{(t)} \end{bmatrix},
\end{equation}
where $[\cdot~\dots~\cdot]$ denotes concatenation.
Here we consider $\beta=2$, enabling to model the positions, velocities, and accelerations. 

\vspace{-3mm}
\subsection{Decoder} 
The decoder targets to predict future poses based on the learned high-level dynamics from the observed motions; see the right part in Fig.~\ref{fig:pipeline}. The core of the decoder is a novel module: \emph{graph-based attention GRU} (GA-GRU). Its functionality is to learn and update hidden states with a trainable graph, as well as to predict the poses based on the motion feature enhanced by an attention mechanism.

A cell of the GA-GRU contains two trainable graphs and a series of embedding networks for state updating. In one GA-GRU cell, let $\mathbf{A}_{\rm I}$ and $\mathbf{A}_{\rm H} \in \mathbb{R}^{M \times M}$ be the adjacency matrices of the inbuilt graphs associated with the input data and hidden features across the joint dimension, which are initialized with the skeleton-graph and trained to build adaptive edges. We use graph convolution with the trainable graphs to capture the attention of each joint. The intuition is that during motion prediction, different joints have different importance on the motion expression. The attention mechanism focuses more on the important joint, increasing the model flexibility to make progress in the improvement of the performance.

At time $T<t<T+\Delta T$, let $\widehat{{\bf X}}^{(t)}$ be the pose predicted at the time step $t-1$, ${\bf I}^{(t)}={\rm diff}_{2}(\widehat{{\bf X}}^{(t)})$ consist of three orders of differences of $\widehat{{\bf X}}^{(t)}$,  and $\mathbf{H}^{(t)}$ be the state matrix of the GA-GRU at time $t-1$. Then the GA-GRU cell works as
\begin{subequations}
\setlength{\abovedisplayskip}{4pt}
\setlength{\belowdisplayskip}{4pt}
    \begin{align}
        {\bf I}_{\rm a}^{(t)} &= \sigma
                                  \left (
                                  {\rm ReLU}
                                  \left (
                                  {\bf A}_{\rm I}{\bf I}^{(t)}{\bf W}_{{\rm I}}
                                  \right )
                                  {\bf U}_{{\rm I}}
                                  \right )
                                  \otimes {\bf I}^{(t)}
                                  \label{eq:I_att}
                                  \\
        {\bf H}_{\rm a}^{(t)} &= \sigma
                                  \left (
                                  {\rm ReLU}
                                  \left (
                                  {\bf A}_{\rm H}{\bf H}^{(t)}{\bf W}_{{\rm H}}
                                  \right )
                                  {\bf U}_{{\rm H}}
                                  \right )
                                  \otimes {\bf H}^{(t)}
                                  \label{eq:H_att}
                                  \\
        \mathbf{r}^{(t)} &= \sigma
                             \left (
                             r_{\rm in}
                             \left(
                             \mathbf{I}_{\rm a}^{(t)}
                             \right ) + 
                            r_{\rm h}
                             \left(
                             {\bf H}_{\rm a}^{(t)}
                             \right )
                             \right )
                             \label{eq:GRU_r}
                             \\
        \mathbf{u}^{(t)} &= \sigma
                             \left(
                             u_{\rm in}
                             \left(
                             \mathbf{I}_{\rm a}^{(t)}
                             \right ) + 
                            u_{\rm h}
                             \left(
                             {\bf H}_{\rm a}^{(t)}
                             \right )
                             \right )
                             \label{eq:GRU_u}
                             \\
        \mathbf{c}^{(t)} &= {\rm tanh}
                             \left (
                             c_{\rm in}
                             \left(
                             \mathbf{I}_{\rm a}^{(t)}
                             \right ) + 
                            \mathbf{r}^{(t)} \odot c_{\rm h}
                             \left(
                             {\bf H}_{\rm a}^{(t)}
                             \right )
                             \right )
                             \label{eq:GRU_c}
                             \\
        \mathbf{H}^{(t+1)} &= \mathbf{u}^{(t)} \odot \mathbf{H}^{(t)} + 
                            \left (
                            1-\mathbf{u}^{(t)}
                            \right ) \odot \mathbf{c}^{(t)},
                            \label{eq:GRU_H}
    \end{align}
\end{subequations}
where $r_{\rm in}(\cdot)$, $r_{\rm h}(\cdot)$, $u_{\rm in}(\cdot)$, $u_{\rm h}(\cdot)$, $c_{\rm in}(\cdot)$ and $c_{\rm h}(\cdot)$ are trainable linear mappings; $\mathbf{W}_{{\rm I}}$, $\mathbf{U}_{{\rm I}}$, $\mathbf{W}_{{\rm H}}$ and $\mathbf{U}_{{\rm H}}$ denote the trainable weights when we learn the attention scores; $\sigma(\cdot)$ is the sigmoid function; $\odot$ is the element-wise product of two vectors, while $\otimes$ is the vector-matrix product at each row with broadcast. In~\eqref{eq:I_att} and~\eqref{eq:H_att}, the graph convolutions achieve information propagation and produces the attention weights to enhance the most important joint features; From~\eqref{eq:GRU_r} to~\eqref{eq:GRU_H}, the GA-GRU updates the motion state.

Compared to standard GRU~\cite{Cho_2014_EMNLP,Martinez_2017_CVPR, Gui_2018_ECCV}, which only uses the vectorized pose features, the proposed GA-GRU exploits the internal spatial relations between different body joints, providing additional information to assist motion prediction. 
Compared to other graph-based GRU~\cite{SymGNN,Li_2020_CVPR}, our GA-GRU does not directly use the motion states processed by graph networks. We calculate an attention map to represent the importance of different pose components to enhance the key components for motion prediction.
Notably, compared to graph attention network~\cite{velickovic2018graph}, which obtains edge attention based on vertex embedding, we here capture the vertex attentions based on both the graph structure and vertex embedding.

We next generate future pose displacements with an output function. Finally, we add the displacements to the input pose to predict the next frame. At frame $t$, the decoder works as
\begin{equation}
\setlength{\abovedisplayskip}{4pt}
\setlength{\belowdisplayskip}{4pt}
    \widehat{\mathbf{X}}^{(t+1)} = \widehat{\mathbf{X}}^{(t)} + f_{\rm pd}\left(\textrm{GA-GRU}\left( {\rm diff}_{2}(\mathbf{\widehat{X}}^{(t)}), \mathbf{H}^{(t)}\right)\right),
\end{equation}
where $f_{\rm pd}(\cdot)$ denotes an readout function, implemented by an MLP. The initial hidden state is the output of the encoder.

\vspace{-3mm}
\subsection{Loss Function}
To train our model, we propose the loss functions.

\textbf{Prediction loss.}
We first introduce a prediction loss, which trains the model to reduce the distances between the predicted samples and ground-truths. Here we consider the $\ell_1$ loss. 
Let a predicted motion be $\widehat{\mathbb{X}}^{+} \in \mathbb{R}^{\Delta T \times M\times 3}$ and the corresponding ground truth be $\mathbb{X}^{+}$, the prediction loss function is
\begin{equation}
\setlength{\abovedisplayskip}{4pt}
\setlength{\belowdisplayskip}{4pt}
\label{eq:pred}
\mathcal{L}_{\rm pred} =  
                          \left \|
                          {\rm vec}
                          \left (
                          \mathbb{X}^{+} - 
                          \widehat{\mathbb{X}}^{+}
                          \right )
                          \right\|_1,
\end{equation}
where $||\cdot||_1$ denotes the $\ell_1$ norm of a vector, and ${\rm vec}(\cdot)$ vectorizes a tensor to a 1D vector. The $\ell_1$ loss gives sufficient gradients to joints with small losses to promote precise prediction; the $\ell_1$ loss also gives stable gradients to joints with large losses, alleviating gradient explosion.

\textbf{Gram matrix loss:}
Inspired by~\cite{ijcai2018-130},
to constrain the predictions to carry the temporal dependencies, we use a gram matrix loss to minimize the distance between the covariances of predicted and ground-truth motions. At time $t$, let the ground-truth position of the $i$th joint be ${\bf x}_i^{(t)}\in\mathbb{R}^{3}$ and the predicted position be $\widehat{{\bf x}}_i^{(t)}\in\mathbb{R}^{3}$, We define the gram matrix of the ground-truth joint positions at two consecutive frames as ${\bf V}_i^{(t-1,t)}=[{\bf x}_i^{(t-1)}~{\bf x}_i^{(t)}][{\bf x}_i^{(t-1)}~{\bf x}_i^{(t)}]^{\top}\in\mathbb{R}^{D_{\bf x} \times D_{\bf x}}$, where $[\cdot~\cdot]$ is concatenation along time; as well as the gram matrix of the predicted joint positions is $\widehat{{\bf V}}_i^{(t-1,t)}=[\widehat{{\bf x}}_i^{(t-1)}~\widehat{{\bf x}}_i^{(t)}][\widehat{{\bf x}}_i^{(t-1)}~\widehat{{\bf x}}_i^{(t)}]^{\top}$. The gram matrix loss is
\begin{equation}
\setlength{\abovedisplayskip}{4pt}
\setlength{\belowdisplayskip}{4pt}
\label{eq:gram}
\mathcal{L}_{\rm gram} =  \frac{1}{\Delta T}
                          \sum_{i=1}^{M}
                          \sum_{t=T+1}^{T+\Delta T}
                          \left \|
                          {\bf V}_i^{(t-1,t)}
                          -
                          \widehat{{\bf V}}_i^{(t-1,t)}
                          \right \|_F^2,
\end{equation}
where $\|\cdot\|_F$ denotes the Frobenius norm of the matrix.
In this way, the model can be effectively trained to preserve the pose changing, as well as alleviate collapsing to a mean pose, especially for the long-term motion prediction.

\textbf{Entropy loss:}
To achieve more decisive grouping of body joints in the multiscale graph generation, we constrain the spatial graph pooling operator by regularizing each row of the clustering matrix to be sparse. Here, we utilize an entropy loss on the assignment for each joint. Given the clustering matrix ${\Psi}_{0 \to s}$, the entropy loss is
\begin{equation}
\setlength{\abovedisplayskip}{4pt}
\setlength{\belowdisplayskip}{4pt}
\label{eq:ent}
\mathcal{L}_{\rm ent} = -\frac{1}{M}
                        \sum_{i=1}^{M}
                        \left ( {\Psi}_{0 \to s}
                        \right )_{i}
                        \left ( \log {\Psi}_{0 \to s}
                        \right )_{i}^{\top},
\end{equation}
where $(\cdot)_{i}$ denotes the $i$th row of a matrix. Since each row of ${\Psi}_{0 \to s}$ has been normalized by a softmax function, which could be regarded as the assignment probability for each cluster, we can directly use it to minimize the entropy.

{
There are two main reasons that we introduce an entropy loss. First, we expect a body component to carry the information of only a few fine-scale nodes to achieve a clear joint clustering.
To this end, we constrain any joint to be clustered in a few even a unique component. Without the entropy loss, the graph pooling tends transform any joint to each cluster in an average and diffuse way, causing the specific information loss across scales; with the entropy loss, each joint could be mapped into a few main clusters with much stronger weights. Theoretically, though each node can automatically find the most suitable cluster through machine learning, our experimental investigation proved that the constraint of entropy loss is necessary. 
Second, the other perspective is from the cross-scale relationships. Cross-scale information is propagated through the learned cross-scale relations. The sparse relations effectively emphasize the dominate association and influence across two scales. With the entropy loss, the coarse nodes could capture the information from the mainly related joints and reduce the information redundancy.
}

Overall, the final loss function for model training is formulated by the weighted sum of the proposed loss; that is,
\begin{equation}
\setlength{\abovedisplayskip}{4pt}
\setlength{\belowdisplayskip}{4pt}
    \mathcal{L} = \alpha \mathcal{L}_{\rm pred} 
    + \beta \mathcal{L}_{\rm gram} 
    + \gamma \mathcal{L}_{\rm ent},
\end{equation}
where the hyper-parameters $\alpha$, $\beta$, and $\gamma$ balance the different terms of loss to train the model for accurate motion prediction.

\section{Experiments}
\label{sec:Experiments}
\subsection{Datasets and Experimental Setups}
\textbf{Dataset 1: Human 3.6M (H3.6M)}
H3.6M dataset~\cite{6682899} has $7$ subjects performing $15$ classes of actions. There are $32$ joints in each subject, and we transform the joint positions into the exponential maps and only use the joints with non-zero values ($20$ joints remain). Along the time axis, we downsample all sequences by two. Following previous paradigms~\cite{Martinez_2017_CVPR}, the models are trained on $6$ subjects and tested on the specific clips of the $5$th subject.

\textbf{Dataset 2: CMU motion capture (CMU Mocap)}
CMU Mocap dataset consists of $5$ general classes of actions, where each subject has $38$ joints and we preserve $26$ joints with non-zero exponential maps. Following~\cite{Li_2018_CVPR}, we use $8$ actions: `basketball', `basketball signal', `directing traffic', `jumping', `running', `soccer', `walking' and `washing window'.

\textbf{Dataset 3: 3D Pose in the Wild (3DPW)}
The 3D Pose in the Wild dataset (3DPW) \cite{Marcard_2018_ECCV} is a large-scale dataset that contains more than 51k frames with 3D poses for challenging indoor and outdoor activities. We adopt the training, test and validation separation suggested by the official setting. The frame rate of the 3D poses is 30Hz.

\textbf{Model configuration}
We implement MST-GNN with PyTorch 1.0 on one NVIDIA Tesla V100 GPU. We set $3$ spatial scales for both H3.6M and CMU Mocap, which contain the original number of joints, $1/2$ and $1/4$ of the original number of joints as vertices. We set $3$ temporal scales for both datasets, which contains the original number of frames, $1/2$ and $1/3$ of the original number of frames. We use $4$ cascaded MST-GCUs, whose feature dimensions are $64$, $64$, $128$ and $256$, respectively. In the decoder, the dimension of the GA-GRU is $256$, and we use a two-layer MLP for pose output. In training, we set the batch size $32$ and clip the gradients to a maximum $\ell_2$-norm of $0.5$; we use Adam with a learning rate $0.0001$. All the hyper-parameters are selected with validation sets.

\textbf{Baseline methods.} 
We compare the proposed MST-GNN with many recent works, which learned motion patterns from pose vectors, e.g. Res-sup.~\cite{Martinez_2017_CVPR}, 
CSM~\cite{Li_2018_CVPR}, 
Traj-GCN~\cite{Mao_2019_ICCV}, DMGNN~\cite{Li_2020_CVPR} and HisRep~\cite{Mao_2020_ECCV}.

\vspace{-3mm}
\subsection{Comparison to State-of-the-Art Methods}
To validate the MST-GNN, we show the quantitative performance for both short-term and long-term motion prediction on Human 3.6M (H3.6M), CMU Mocap and 3DPW. We also illustrate the predicted samples for qualitative evaluation.

\begin{table*}[t]
    \centering
    \caption{Mean angle errors (MAE) of different methods for short-term prediction on all the $15$ actions of H3.6M. We also present the average prediction results across all the actions. We additionally present several degraded MST-GNN variants: MST-GNN ($R=1$) uses only one scale of spatial and temporal graphs ($R=1$); MST-GNN (fixed ${\bf A}_r$) uses fixed graph to represent spatio-temporal relations; MST-GNN (w/GRU) uses a common GRU in the decoder instead of the proposed GA-GRU.}
    \footnotesize
    \renewcommand{\arraystretch}{1.0}
    \resizebox{1\textwidth}{!}{

        \begin{tabular}{|c|cccc|cccc|cccc|cccc|cccc|cccc|}
        \hline
        Motion & \multicolumn{4}{c|}{Walking} & \multicolumn{4}{c|}{Eating} & \multicolumn{4}{c|}{Smoking} 
               & \multicolumn{4}{c|}{Discussion} & \multicolumn{4}{c|}{Directions} & \multicolumn{4}{c|}{Greeting}\\
        \hline
        millisecond & 80&160&320&400 & 80&160&320&400 & 80&160&320&400 & 80&160&320&400 & 80&160&320&400 & 80&160&320&400 \\
        \hline
        Res-sup~\cite{Martinez_2017_CVPR} & 0.27 & 0.46 & 0.67 & 0.75 & 0.23 & 0.37 & 0.59 & 0.73 & 0.32 & 0.59 & 1.01 & 1.10 & 0.30 & 0.67 & 0.98 & 1.06 & 0.41 & 0.64 & 0.80 & 0.92 & 0.57 & 0.83 & 1.45 & 1.60 \\
        CSM~\cite{Li_2018_CVPR} & 0.33 & 0.54 & 0.68 & 0.73 & 0.22 & 0.36 & 0.58 & 0.71 & 0.26 & 0.49 & 0.96 & 0.92 & 0.32 & 0.67 & 0.94 & 1.01 & 0.39 & 0.60 & 0.80 & 0.91 & 0.51 & 0.82 & 1.21 & 1.38 \\
        Traj-GCN~\cite{Mao_2019_ICCV} & {\bf 0.18} & {0.32} & { 0.49} & {0.56} & {0.17} & 0.31 & 0.52 & 0.62 & 0.22 & 0.41 & 0.84 & 0.79 & {\bf 0.20} & {\bf 0.51} & {0.79} & {\bf 0.86} & {0.26} & {0.45} & 0.70 & 0.79 & {0.35} & {0.61} & {0.96} & {1.13} \\
        DMGNN~\cite{Li_2020_CVPR} & {\bf 0.18} & {\bf 0.31} & 0.49 & 0.58 & 0.17 & 0.30 & 0.49 & 0.59 & {\bf 0.21} & 0.40 & 0.81 & 0.78 & 0.26 & 0.65 & 0.92 & 0.99 & 0.25 & 0.44 & 0.65 & 0.71 & 0.36 & 0.61 & 0.94 & 1.12 \\
        Hisrep~\cite{Mao_2020_ECCV} & {\bf 0.18} & {\bf 0.30} & {\bf 0.46} & {\bf 0.51} & {\bf 0.16} & 0.29 & 0.49 & 0.60 & 0.22 & 0.42 & 0.86 & 0.80 & {\bf 0.20} & 0.52 & {\bf 0.78} & 0.87 & 0.25 & {\bf 0.43} & {\bf 0.60} & 0.69 & 0.35 & 0.60 & {\bf 0.95} & 1.14 \\
        \hline
        MST-GNN ($R=1$) & 0.20 & 0.34 & 0.52 & 0.61 & 0.19 & 0.32 & 0.52 & 0.64 & 0.23 & 0.42 & 0.82 & 0.80 & 0.24 & 0.62 & 0.88 & 0.96 & 0.25 & 0.45 & 0.64 & 0.69 & 0.35 & {\bf 0.58} & 0.97 & 1.16 \\
        MST-GNN (fixed ${\bf A}_r$) & 0.20 & 0.34 & 0.54 & 0.60 & 0.20 & 0.34 & 0.54 & 0.65 & 0.24 & 0.42 & 0.84 & 0.81 & 0.26 & 0.65 & 0.91 & 1.00 &0.27 & 0.46 & 0.64 & 0.72 & 0.37 & 0.59 & 0.97 & 1.18  \\
        MST-GNN (w/GRU) & 0.21 & 0.33 & 0.52 & 0.62 & 0.19 & 0.33 & 0.54 & 0.65 & 0.24 & 0.42 & 0.86 & 0.84 & 0.26 & 0.61 & 0.87 & 0.94 & 0.25 & 0.46 & 0.63 & 0.71 & 0.35 & 0.59 & 0.96 & 1.16 \\
        MST-GNN & {\bf 0.18} & {0.31} & {0.49} & {0.57} & {\bf 0.16} & {\bf 0.28} & {\bf 0.47} & {\bf 0.55} & {\bf 0.21} & {\bf 0.39} & {\bf 0.78} & {\bf 0.77} & {0.22} & {0.56} & {0.83} & {0.89} & {\bf 0.24} & {\bf 0.43} & {\bf 0.60} & {\bf 0.67} & {\bf 0.34} & {\bf 0.58} & {\bf 0.95} & {\bf 1.12} \\
        \hline
        Motion & \multicolumn{4}{c|}{Phoning} & \multicolumn{4}{c|}{Posing} & \multicolumn{4}{c|}{Purchases} 
               & \multicolumn{4}{c|}{Sitting} & \multicolumn{4}{c|}{Sitting Down} & \multicolumn{4}{c|}{Taking Photo} \\
        \hline
        millisecond & 80&160&320&400 & 80&160&320&400 & 80&160&320&400 & 80&160&320&400 & 80&160&320&400 & 80&160&320&400 \\
        \hline
        Res-sup~\cite{Martinez_2017_CVPR} & 0.59 & 1.06 & 1.45 & 1.60 & 0.45 & 0.85 & 1.34 & 1.56 & 0.58 & 0.79 & 1.08 & 1.15 & 0.41 & 0.68 & 1.12 & 1.33 & 0.47 & 0.88 & 1.37 & 1.54 & 0.28 & 0.57 & 0.90 & 1.02 \\
        CSM~\cite{Li_2018_CVPR} & 0.59 & 1.13 & 1.51 & 1.65 & 0.29 & 0.60 & 1.12 & 1.37 & 0.63 & 0.91 & 1.19 & 1.29 & 0.39 & 0.61 & 1.02 & 1.18 & 0.41 & 0.78 & 1.16 & 1.31 & 0.23 & 0.49 & 0.88 & 1.06 \\
        Traj-GCN~\cite{Mao_2019_ICCV} & {0.53} & {1.02} & 1.32 & 1.45 & 0.23 & 0.54 & 1.26 & 1.38 & {0.42} & 0.66 & 1.04 & 1.12 & 0.29 & 0.45 & 0.82 & {0.97} & {\bf 0.30} & {0.63} & {0.89} & {1.01} & {\bf 0.15} & {0.36} & {0.59} & {0.72} \\
        DMGNN~\cite{Li_2020_CVPR} & 0.52 & 0.97 & 1.29 & 1.43 & 0.20 & 0.46 & 1.06 & 1.34 & 0.41 & 0.61 & 1.05 & 1.14 & {\bf 0.26} & 0.42 & 0.76 & 0.97 & 0.32 & 0.65 & 0.93 & 1.05 & {\bf 0.15} & {\bf 0.34} & 0.58 & 0.71 \\
        Hisrep~\cite{Mao_2020_ECCV} & 0.53 & 1.01 & 1.31 & 1.43 & 0.19 & 0.46 & 1.09 & 1.35 & 0.42 & 0.65 & 1.00 & 1.07 & 0.29 & 0.47 & 0.83 & 1.01 & {\bf 0.30} & 0.63 & 0.92 & 1.04 & 0.16 & 0.36 & 0.58 & 0.70 \\
        \hline
        MST-GNN ($R=1$) & {\bf 0.52} & 0.87 & 1.26 & 1.39 & 0.19 & 0.46 & 1.05 & 1.27 & {\bf 0.40} & 0.61 & 0.99 & 1.06 & {\bf 0.26} & 0.42 & 0.76 & 0.93 & 0.31 & 0.63 & 0.92 & 1.02 & {\bf 0.15} & {\bf 0.34} & 0.58 & {\bf 0.69} \\
        MST-GNN (fixed ${\bf A}_r$) & 0.54 & 0.92 & 1.28 & 1.42 & 0.20 & 0.46 & 1.05 & 1.25 & 0.42 & 0.64 & 1.02 & 1.07 & 0.29 & 0.44 & 0.79 & 0.97 & 0.33 & 0.64 & 0.95 & 1.03 & 0.17 & 0.37 & 0.59 & 0.72 \\
        MST-GNN (w/GRU) & 0.53 & 0.88 & 1.26 & 1.40 & 0.20 & {\bf 0.44} & 1.02 & 1.22 & 0.41 & 0.62 & 0.99 & {\bf 1.04} & 0.27 & 0.44 & 0.77 & 0.96 & 0.32 & 0.63 & 0.93 & 1.03 & 0.17 & {\bf 0.34} & 0.59 & 0.70 \\
        MST-GNN & {\bf 0.52} & {\bf 0.83} & {\bf 1.25} & {\bf 1.38} & {\bf 0.18} & {\bf 0.44} & {\bf 0.98} & {\bf 1.20} & {\bf 0.40} & {\bf 0.60} & {\bf 0.97} & {\bf 1.04} & {\bf 0.26} & {\bf 0.41} & {\bf 0.75} & {\bf 0.92} & {\bf 0.30} & {\bf 0.62} & {\bf 0.88} & {\bf 0.99} & {\bf 0.15} & {0.35} & {\bf 0.57} & {\bf 0.69} \\
        \hline
        Motion & \multicolumn{4}{c|}{Waiting} & \multicolumn{4}{c|}{Walking Dog} & \multicolumn{4}{c|}{Walking Together} & \multicolumn{4}{c|}{Average}\\
        \cline{1-17}
        millisecond & 80&160&320&400 & 80&160&320&400 & 80&160&320&400 & 80&160&320&400 \\
        \cline{1-17}
        Res-sup.~\cite{Martinez_2017_CVPR} & 0.32 & 0.63 & 1.07 & 1.26 & 0.52 & 0.89 & 1.25 & 1.40 & 0.27 & 0.53 & 0.74 & 0.79 & 0.40 & 0.69 & 1.04 & 1.18 \\
        CSM~\cite{Li_2018_CVPR} & 0.30 & 0.62 & 1.09 & 1.30 & 0.59 & 1.00 & 1.32 & 1.44 & 0.27 & 0.52 & 0.71 & 0.74 & 0.38 & 0.68 & 1.01 & 1.13 \\
        Traj-GCN~\cite{Mao_2019_ICCV} & 0.23 & 0.50 & 0.92 & 1.15 & 0.46 & 0.80 & {1.12} & {1.30} & {0.15} & 0.35 & 0.52 & {0.57} & 0.27 & 0.53 & 0.85 & 0.96 \\
        DMGNN~\cite{Li_2020_CVPR} & 0.22 & {\bf 0.49} & {\bf 0.88} & 1.10 & 0.42 & {\bf 0.72} & 1.16 & 1.34 & 0.15 & 0.33 & 0.52 & 0.57 & 0.27 & 0.52 & 0.83 & 0.95 \\
        Hisrep~\cite{Mao_2020_ECCV} & 0.22 & {\bf 0.49} & 0.92 & 1.14 & 0.46 & 0.78 & 1.05 & {\bf 1.23} & {\bf 0.14} & {\bf 0.32} & {\bf 0.50} & {\bf 0.55} & 0.27 & 0.52 & 0.82 & 0.94 \\
        \cline{1-17}
        MST-GNN ($R=1$) & 0.22 & 0.50 & 0.91 & 1.11 & 0.43 & 0.76 & 1.12 & 1.27 & 0.16 & 0.34 & 0.52 & 0.58 & 0.27 & 0.51 & 0.83 & 0.94 \\
        MST-GNN (fixed ${\bf A}_r$) & 0.25 & 0.52 & 0.94 & 1.15 & 0.46 & 0.79 & 1.15 & 1.31 & 0.16 & 0.37 & 0.53 & 0.61 & 0.29 & 0.53 & 0.84 & 0.96 \\
        MST-GNN (w/GRU) & 0.24 & 0.52 & 0.95 & 1.15 & 0.44 & 0.78 & 1.16 & 1.30 & 0.16 & 0.36 & 0.54 & 0.61 & 0.27 & 0.52 & 0.84 & 0.95\\
        MST-GNN & {\bf 0.21} & {\bf 0.49} & {\bf 0.88} & {\bf 1.08} & {\bf 0.41} & {0.73} & {\bf 1.11} & {1.25} & {0.15} & {\bf 0.32} & {0.51} & {\bf 0.55} & {\bf 0.26} & {\bf 0.50} & {\bf 0.79} & {\bf 0.92} \\
        \cline{1-17}
        \end{tabular}}
        \vspace{-3mm}
    \label{tab:pred_h36m_15}
\end{table*}

\textbf{Short-term motion prediction.}
Short-term motion prediction aims to predict the future poses within $400$ milliseconds. We test our model on the $15$ actions of H3.6M dataset. {Table~\ref{tab:pred_h36m_15} shows MAEs various methods on each action as well as the average MAE across $15$ actions.} Besides previous baselines, we also present three MST-GNN variants: 1) we use only the original scale of spatio-temporal graphs ($R=1$); 2) We fix the spatio-temporal graph at any scale $r$ (fixed ${\bf A}_r$); 3) We replace our GA-GRU to the common GRU~\cite{Cho_2014_EMNLP} (w/GRU).
We see that i) the complete MST-GNN obtains more effective prediction than the variants; ii) compared to baselines, MST-GNN has the lowest MAEs on most actions, and obtains competitive results on `Walking' and `Discussion'.

\textbf{Long-term motion prediction.}
Long-term motion prediction aims to predict the poses over $400$ milliseconds, which is challenging due to the action variation. {Table~\ref{tab:longterm} presents the MAEs of various models for predicting $15$ actions and average MAEs across actions in the future 560 ms and 1000 ms on the H3.6M.}
\begin{table*}[t]
    \centering
    \caption{MAEs of methods for long-term prediction on $15$ actions of H3.6M. The average results over all the actions are presented.}
    \renewcommand{\arraystretch}{1.0}
    \footnotesize
    \resizebox{0.92\textwidth}{!}{
    \begin{tabular}{|c|cc|cc|cc|cc|cc|cc|cc|cc|}
        \hline
        Motion & 
        \multicolumn{2}{c|}{Walking} &   \multicolumn{2}{c|}{Eating}&
        \multicolumn{2}{c|}{Smoking} &   \multicolumn{2}{c|}{Discussion} &
        \multicolumn{2}{c|}{Directions} & \multicolumn{2}{c|}{Greeting} &
        \multicolumn{2}{c|}{Phoning} &    \multicolumn{2}{c|}{Posing} \\
        \hline
         
        milliseconds & 560 & 1k & 560 & 1k & 560 & 1k & 560 & 1k & 560 & 1k & 560 & 1k & 560 & 1k & 560 & 1k \\
        \hline
        Res-sup.~\cite{Martinez_2017_CVPR} & 0.93 & 1.03  & 0.95 & {\bf 1.08} & 1.25 & 1.60 & 1.43 & 1.69 & 1.26 & 1.48 & 1.81 & 1.96 & 1.75 & 2.01 & 2.03 & 2.55 \\
        CSM~\cite{Li_2018_CVPR} & 0.88 & 0.92 & 1.01 & 1.24 & 0.97 & 1.62 & 1.56 & 1.86 & 0.98 & 1.37 & 1.72 & 1.79 & 1.68 & 1.92 & 1.93 & 2.62 \\ 
        TP-RNN~\cite{Hsukuang18} & 0.74 & 0.77 & 0.84 & 1.14 & 0.98 & 1.66 & 1.39 & 1.74 & 0.95 & 1.38 & 1.72 & 1.81 &  1.47 & 1.68 & 1.75 & 2.47 \\
        Traj-GCN~\cite{Mao_2019_ICCV} & {0.65} & {0.67} & {0.76} & 1.12 & {0.87} & 1.57 & 1.33 & 1.70 & 0.87 & 1.29 & 1.54 & 1.59 & 1.47 & 1.66 & 1.57 & 2.37 \\ 
        DMGNN~\cite{Li_2020_CVPR} & 0.66 & 0.75 & 0.74 & 1.14 & 0.84 & 1.52 & 1.33 & 1.45 & 0.86 & 1.30 & 1.57 & 1.63 & 1.44 & 1.64 & {\bf 1.49} & 2.17 \\ 
        HisRep~\cite{Mao_2020_ECCV} & 0.59 & {\bf 0.64} & {\bf 0.74} & 1.10 & 0.86 & 1.58 & {\bf 1.29} & 1.63 & {\bf 0.81} & 1.27 & {\bf 1.47} & {\bf 1.57} & 1.41 & 1.68 & 1.60 & 2.32 \\
        \hline
        MST-GNN & {\bf 0.62} & 0.73  & {\bf 0.74} & {1.13} & {\bf 0.83} & {\bf 1.51} & {1.30} & {\bf 1.58} & 0.82 & {\bf 1.26} & 1.55 & 1.61 & {\bf 1.36} & {\bf 1.60} & {1.51} & {\bf 2.15} \\ 
        \hline
        Motion & 
        \multicolumn{2}{c|}{Purchases}& \multicolumn{2}{c|}{Sitting}&
        \multicolumn{2}{c|}{Sitting Down}& \multicolumn{2}{c|}{Taking Photo} &
        \multicolumn{2}{c|}{Waiting} & \multicolumn{2}{c|}{Walking Dog} &
        \multicolumn{2}{c|}{Walking Toge} & \multicolumn{2}{c|}{Average} \\ 
        \hline
        milliseconds & 560 & 1k & 560 & 1k & 560 & 1k & 560 & 1k & 560 & 1k & 560 & 1k & 560 & 1k & 560 & 1k \\
        \hline
        Res-sup.~\cite{Martinez_2017_CVPR} & 1.76 & 2.52 & 1.54 & 1.85 & 1.60 & 2.17 & 1.36 & 1.59 & 1.73 & 2.43 & 1.82 & 2.36 & 1.03 & 1.48 & 1.48 & 1.75 \\
        CSM~\cite{Li_2018_CVPR} & 1.64 & 2.42 & 1.32 & 1.68 & 1.48 & 1.98 & 1.05 & 1.25 & 1.65 & 2.39 & 1.70 & 2.01 & 0.87 & 1.32 & 1.36 & 1.76 \\ 
        TP-RNN~\cite{Hsukuang18} & 1.52 & 2.28 & 1.35 & 1.74 & 1.47 & 1.93 & 1.08 & 1.35 & 1.71 & 2.46 & 1.73 & 1.98 & 0.78 & 1.28 & 1.30 & 1.71 \\
        Traj-GCN~\cite{Mao_2019_ICCV} & 1.46 & 2.26 & 1.15 & 1.50 & 1.20 & 1.72 & 0.86 & 1.08 & 1.58 & 2.32 & {\bf 1.55} & 1.79 & {\bf 0.61} & 1.17 & 1.17 & 1.59 \\ 
        DMGNN~\cite{Li_2020_CVPR} & 1.39 & 2.13 & 1.12 & 1.51 & 1.30 & 1.74 & 0.83 & 1.06 & 1.46 & 2.12 & 1.57 & 1.75 & 0.70 & 1.24 & 1.17 & 1.57 \\ 
        HisRep~\cite{Mao_2020_ECCV} & 1.43 & 2.22 & 1.16 & 1.55 & {\bf 1.18} & {\bf 1.70} & {\bf 0.82} & 1.08 & 1.54 & 2.30 & 1.57 & 1.82 & 0.63 & {\bf 1.16} & {\bf 1.14} & 1.57 \\
        \hline
        MST-GNN & {\bf 1.34} & {\bf 2.04} & {\bf 1.09} & {\bf 1.47} & 1.23 & 1.72 & {\bf 0.82} & {\bf 1.03} & {\bf 1.40} & {\bf 2.05} & 1.56 & {\bf 1.74} & 0.72 & 1.23 & 1.15 & {\bf 1.55} \\ 
        \hline
    \end{tabular}}
    \vspace{-3mm}
    \label{tab:longterm}
\end{table*}
{We see that, MST-GNN achieves more precise prediction on many actions. For example, for smoking, we outperforms baselines by $4.13\%$; for phoning, we outperforms baselines by $4.34\%$; and for purchases, we outperforms baselines by $7.77\%$, Also, MST-GNN obtains comparable average MAEs across all the actions.}

We also test our MST-GNN for short-term and long-term prediction on CMU Mocap dataset.  Table~\ref{tab:pred_cmu} shows the MAEs of each action and the average results within the future 1000 ms. We see that MST-GNN significantly outperforms the state-of-the-art methods by {$11.84\%$} and {$4.71\%$} for short-term and long-term prediction, respectively.
\begin{table*}[t]
    \centering
    \caption{Prediction MAEs of our MST-GNN and the state-of-the-art methods on the 8 actions of CMU Mocap for both short-term and long-term motion prediction. We also present the average prediction results across all the actions.}
    \vspace{-1mm}
    \footnotesize
    \renewcommand{\arraystretch}{1.0}
    \resizebox{0.8\textwidth}{!}{
        \begin{tabular}{|c|cccc|c|cccc|c|cccc|c|}
        \hline
        Motion & \multicolumn{5}{c|}{Basketball} & \multicolumn{5}{c|}{Basketball Signal} & \multicolumn{5}{c|}{Directing Traffic}  \\ \hline
        milliseconds & 80 & 160 & 320 & 400 & 1000 & 80 & 160 & 320 & 400 & 1000 & 80 & 160 & 320 & 400 & 1000 \\ \hline
        Res-sup.~\cite{Martinez_2017_CVPR} & 0.49 & 0.77 & 1.26 & 1.45 & 1.77 & 0.42 & 0.76 & 1.33 & 1.54 & 2.17 & 0.31 & 0.58 & 0.94 & 1.10 & 2.06 \\
        CSM~\cite{Li_2018_CVPR} & 0.37 & 0.62 & 1.07 & 1.18 & 1.95 & 0.32 & 0.59 & 1.04 & 1.24 & 1.96 & 0.25 & 0.56 & 0.89 & 1.00 & 2.04 \\
        Traj-GCN~\cite{Mao_2019_ICCV} & 0.33 & 0.52 & {0.89} & {1.01} & 1.71 & 0.11 & 0.20 & 0.41  & 0.53 & {\bf 1.00} & {\bf 0.15} & {0.32} & {0.52} & {\bf 0.60} & 2.00 \\
        DMGNN~\cite{Li_2020_CVPR} & 0.33 & 0.46 & 0.89 & 1.11 & {\bf 1.66} & 0.10 & {\bf 0.17} & 0.31 & 0.41 & 1.26 & {\bf 0.15} & 0.30 & 0.57 & 0.72 & 1.98 \\
        \hline
        MST-GNN & {\bf 0.28} & {\bf 0.49} & {\bf 0.88} & {\bf 1.01} & {1.68} & {\bf 0.10} & {0.18} & {\bf 0.29} & {\bf 0.38} & 1.04 & {\bf 0.15} & {\bf 0.29} & {\bf 0.51} & {0.62} & {\bf 1.95} \\ 
        \hline
        Motion & \multicolumn{5}{c|}{Jumping} & \multicolumn{5}{c|}{Running} & \multicolumn{5}{c|}{Soccer} \\ \hline
        milliseconds & 80 & 160 & 320 & 400 & 1000 & 80 & 160 & 320 & 400 & 1000 & 80 & 160 & 320 & 400 & 1000 \\ \hline
        Res-sup.~\cite{Martinez_2017_CVPR} & 0.57 & 0.86 & 1.76 & 2.03 & 2.42 & 0.32 & 0.48 & 0.65 & 0.74 & 1.00 & 0.29 & 0.50 & 0.87 & 0.98 & 1.73 \\
        CSM~\cite{Li_2018_CVPR} & 0.39 & 0.60 & {1.36} & {1.56} & 2.01 & 0.28 & 0.41 & {0.52} & 0.57 & {0.67} & 0.26 & 0.44 & 0.75 & 0.87 & 1.56 \\
        Traj-GCN~\cite{Mao_2019_ICCV} & {\bf 0.31} & {\bf 0.49} & {\bf 1.23} & {\bf 1.39} & {1.80} & 0.33 & 0.55 & 0.73 & 0.74 & 0.95 & {\bf 0.18} & {0.29} & {\bf 0.61}  & {\bf 0.71} & 1.40 \\
        DMGNN~\cite{Li_2020_CVPR} & 0.37 & 0.65 & 1.49 & 1.71 & {\bf 1.79} & 0.19 & 0.31 & 0.47 & 0.49 & {\bf 0.64} & 0.22 & 0.32 & 0.79 & 0.91 & 1.54 \\
        \hline
        MST-GNN & {\bf 0.31} & {0.53} & 1.29 & 1.46 & {1.82} & {\bf 0.18} & {\bf 0.28} & {\bf 0.43} & {\bf 0.48} & {0.66} & {\bf 0.18} & {\bf 0.28} & {0.63} & {\bf 0.71} & {\bf 1.38}\\ 
        \hline
        Motion & \multicolumn{5}{c|}{Walking} & \multicolumn{5}{c|}{Washing Window} & \multicolumn{5}{c|}{Average} \\
        \hline
        milliseconds & 80 & 160 & 320 & 400 & 1000 & 80 & 160 & 320 & 400 & 1000 & 80 & 160 & 320 & 400 & 1000 \\
        \hline
        Res-sup.~\cite{Martinez_2017_CVPR} & 0.35 & 0.45 & 0.59 & 0.64 & 0.88 & 0.32 & 0.47 & 0.74 & 0.93 & 1.37 & 0.38 & 0.61 & 1.02 & 1.18 & 1.68 \\
        CSM~\cite{Li_2018_CVPR} & 0.35 & 0.44 & 0.45 & 0.50 & 0.78 & 0.30 & 0.47 & 0.80 & 1.01 & 1.39 & 0.32 & 0.52 & 0.88 & 0.99 & 1.55 \\
        Traj-GCN~\cite{Mao_2019_ICCV} & 0.33 & 0.45 & 0.49 & 0.53 & 0.61 & 0.22 & 0.33 & 0.57 & {0.75} & 1.20 & 0.24 & 0.39 & 0.68 & 0.78 & 1.33 \\
        DMGNN~\cite{Li_2020_CVPR} & 0.30 & 0.34 & 0.38 & 0.43 & 0.60 & {\bf 0.20} & {\bf 0.27} & 0.62 & 0.81 & 1.09 & 0.23 & 0.35 & 0.69 & 0.82 & 1.32 \\
        \hline
        MST-GNN & {\bf 0.26} & {\bf 0.32} & {\bf 0.37} & {\bf 0.41} & {\bf 0.54} & {0.21} & {0.31} & {\bf 0.54} & {\bf 0.69} & {\bf 1.07} & {\bf 0.21} & {\bf 0.33} & {\bf 0.62} & {\bf 0.72} & {\bf 1.27} \\
        \hline
    \end{tabular}}
    \vspace{-2mm}
    \label{tab:pred_cmu}
\end{table*}

We further test the MST-GNN on 3DPW dataset for both short-term and long-term prediction. We present the average MAEs across test samples in Table~\ref{tab:MAE_3DPW}.
  \begin{table}[t]
      \small 
      \centering
      \caption{Average MAEs of methods on 3DPW dataset at various prediction time steps.}
      \vspace{-2mm}
      \begin{tabular}{|c|ccccc|}
          \hline
          ~ & \multicolumn{5}{c|}{Average MAE} \\
          \hline
          milliseconds & 200 & 400 & 600 & 800 & 1000 \\
          \hline
          Res-sup.~\cite{Martinez_2017_CVPR} & 1.95 & 2.37 & 2.46 & 2.51 & 2.53 \\
          CSM~\cite{Li_2018_CVPR} & 1.24 & 1.85 & 2.13 & 2.23 & 2.26  \\
          Traj-GCN~\cite{Mao_2019_ICCV} & 0.64 & 0.94 & {\bf 1.11} & 1.22 & 1.27 \\
          DMGNN~\cite{Li_2020_CVPR} & 0.62 & 0.93 & 1.14 & 1.23 & 1.26 \\
          HisRep~\cite{Mao_2020_ECCV} & 0.64 & 0.96 & 1.16 & {\bf 1.21} & {\bf 1.22} \\
          \hline
          MST-GNN & {\bf 0.61} & {\bf 0.91} & {1.13} & {\bf 1.21} & 1.24 \\
          \hline
      \end{tabular}
      \vspace{-5mm}
      \label{tab:MAE_3DPW}
  \end{table}
  Compared to the state-of-the-art methods, at very short terms (e.g., 80 ms and 160 ms), MST-GNN significantly outperforms previous works. For long-term prediction, our MST-GNN also achieves more effective prediction than other methods by $1.13\%$.
  
\begin{figure*}[t]
  \begin{center}
    \begin{tabular}{cc}
     \includegraphics[width=0.98\columnwidth]{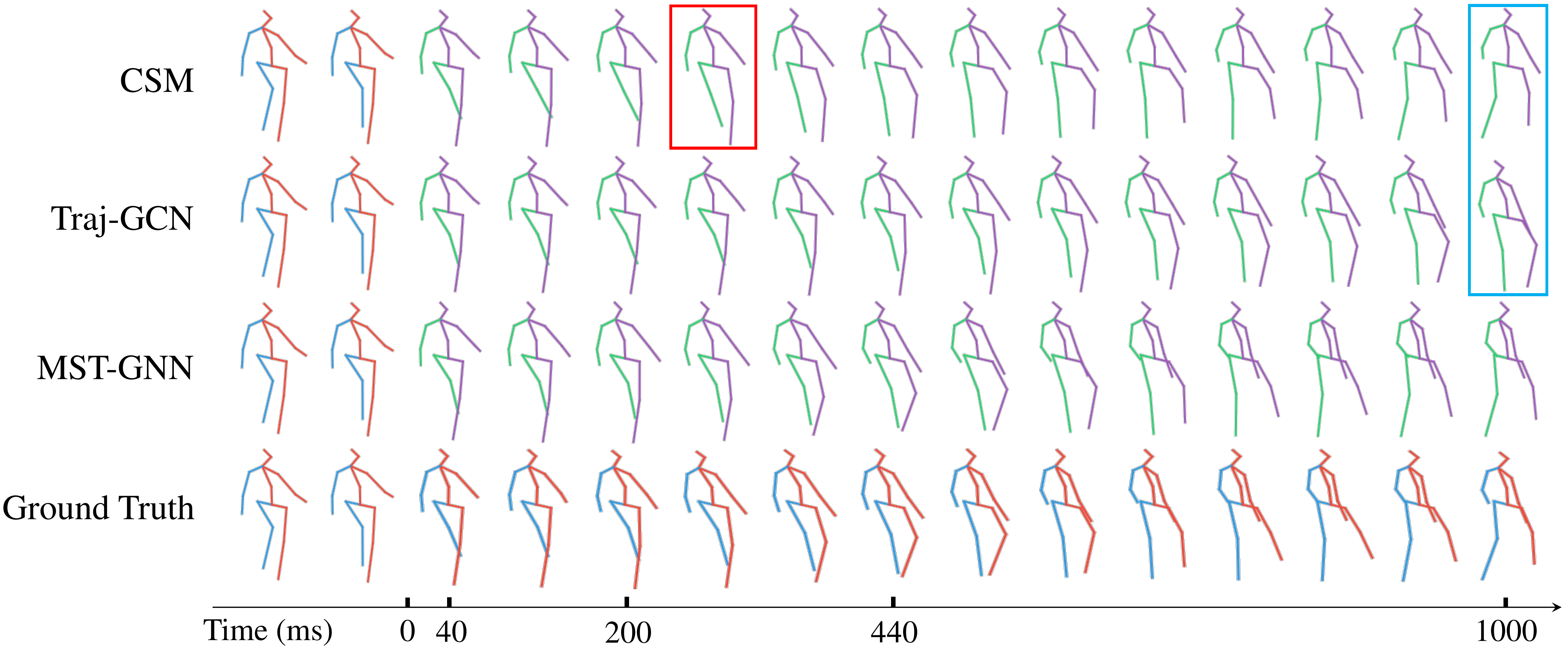} & 
     \includegraphics[width=0.98\columnwidth]{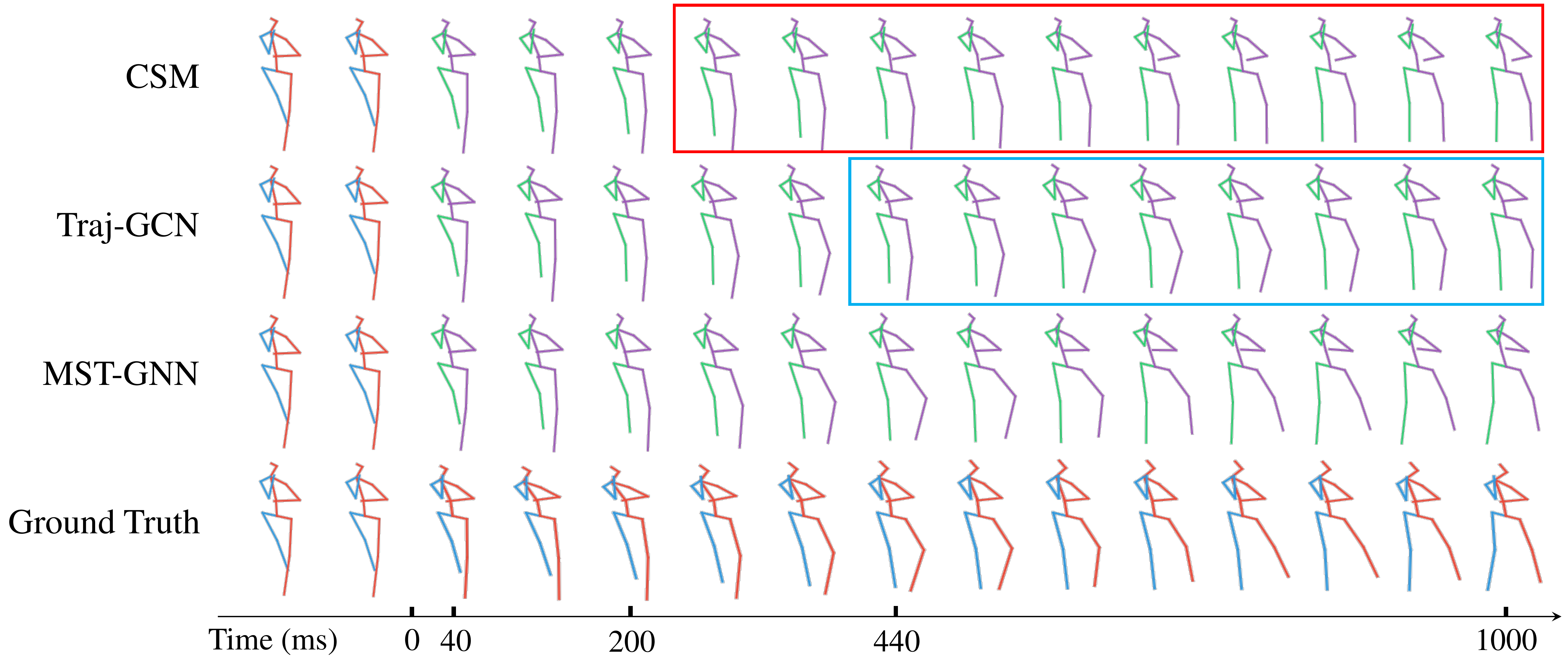}
     \\
     {{\small (a) Action of walking.}} & 
     {{\small (b) Action of smoking.}}
  \end{tabular}
\end{center}
\vspace{-10pt}
\caption{\small Qualitative comparison on two actions in H3.6M for short and long-term prediction. (a) Action of 'Walking'; (b) Action of 'Smoking'.}
\vspace{-3mm}
\label{fig:sample_show}
\end{figure*}

\textbf{Predicted sample visualization.} We compare the synthesized samples of MST-GNN to those of CSM and Traj-GCN on H3.6M. Fig.~\ref{fig:sample_show} illustrates the future poses of `Walking' (a) and `Smoking' (b) in 1000 ms with a frame interval of 80 ms. Compared to baselines, MST-GNN completes the action accurately and reasonably, providing significantly better predictions. In Fig.~\ref{fig:sample_show} (a), the predictions of CSM start to show large errors at the 280th ms (red box); CSM and Traj-GCN have large errors for long-term prediction (blue box); in Fig.~\ref{fig:sample_show} (b), the predictions of two baselines tend to converge to mean poses in long-term (red and blue boxes).

\vspace{-2mm}
\subsection{Ablation Study}
Here we investigate the properties of different model components in our MST-GCN.

\textbf{Effects of multiple scales of the spatial and temporal graphs.}
We first test various scales of spatial and temporal graphs to verify the multiscale graph representation.
\begin{table}[t]
    \centering
    \caption{MAEs of MST-GNN with various spatial scales (noted by vertex numbers) for short-term prediction.}
    \vspace{-2mm}
    \footnotesize
    \renewcommand{\arraystretch}{1.0}
    \resizebox{1\columnwidth}{!}{
        \centering
        \begin{tabular}{|ccccc|ccccc|}
        \hline
         \multicolumn{5}{|c|}{Spatial scale} & \multicolumn{5}{|c|}{MAEs} \\
        \hline
         20 & 10 & 5 & 3 & 2 & 80 & 160 & 320 & 400 & Avg.\\
        \hline
          \checkmark &  &  &  &  & 0.271 & 0.515 & 0.833 & 0.940 & 0.605 \\
          \checkmark & \checkmark &  &  &  & 0.264 & 0.502 & 0.815 & 0.926 & 0.591 \\
          \checkmark &  & \checkmark &  &  & 0.263 & 0.499 & 0.819 & 0.931 & 0.593 \\
          \checkmark &  &  & \checkmark &  & 0.266 & 0.508 & 0.820 & 0.935 & 0.596 \\
          \checkmark &  &  &  & \checkmark & 0.268 & 0.508 & 0.817 & 0.935 & 0.595 \\
          \checkmark & \checkmark & \checkmark &  &  & {\bf 0.261} & {\bf 0.496} & {\bf 0.794} & {\bf 0.919} & {\bf 0.589} \\
          \checkmark & \checkmark &  & \checkmark &  & 0.263 & 0.501 & 0.816 & 0.925 & 0.592 \\
          \checkmark & \checkmark &  &  & \checkmark & 0.264 & 0.503 & {0.814} & 0.928 & 0.594\\
          \checkmark & \checkmark & \checkmark & \checkmark &  & {\bf 0.261} & {\bf 0.496} & 0.815 & 0.921 & 0.590 \\
          \checkmark & \checkmark & \checkmark &  & \checkmark & 0.263 & 0.499 & 0.818 & 0.922 & 0.592\\
          \checkmark & \checkmark & \checkmark & \checkmark & \checkmark & 0.262 & 0.496 & 0.818 & 0.921 & 0.590 \\
        \hline
        \end{tabular}}
    \label{tab:SpatialScaleMAE}
    \vspace{-5mm}
\end{table}
For the multiscale spatial graphs, besides the three spatial scales set in our model, we introduce another two scales that have 3 and 2 vertices to depict much coarser graphs, where we compress the graphs to obtain more abstract information. With various spatial scales, Table~\ref{tab:SpatialScaleMAE} presents the MAEs of short-term prediction at several prediction steps and the mean MAEs over 400 ms, where the used scales are noted by checkmarks. We see that, when we combine the scales with 20, 10, and 5 vertices, MST-GNN obtains the lowest MAEs. 

\begin{figure*}[t]
  \begin{center}
    \begin{tabular}{cccc}
     \includegraphics[width=0.47\columnwidth]{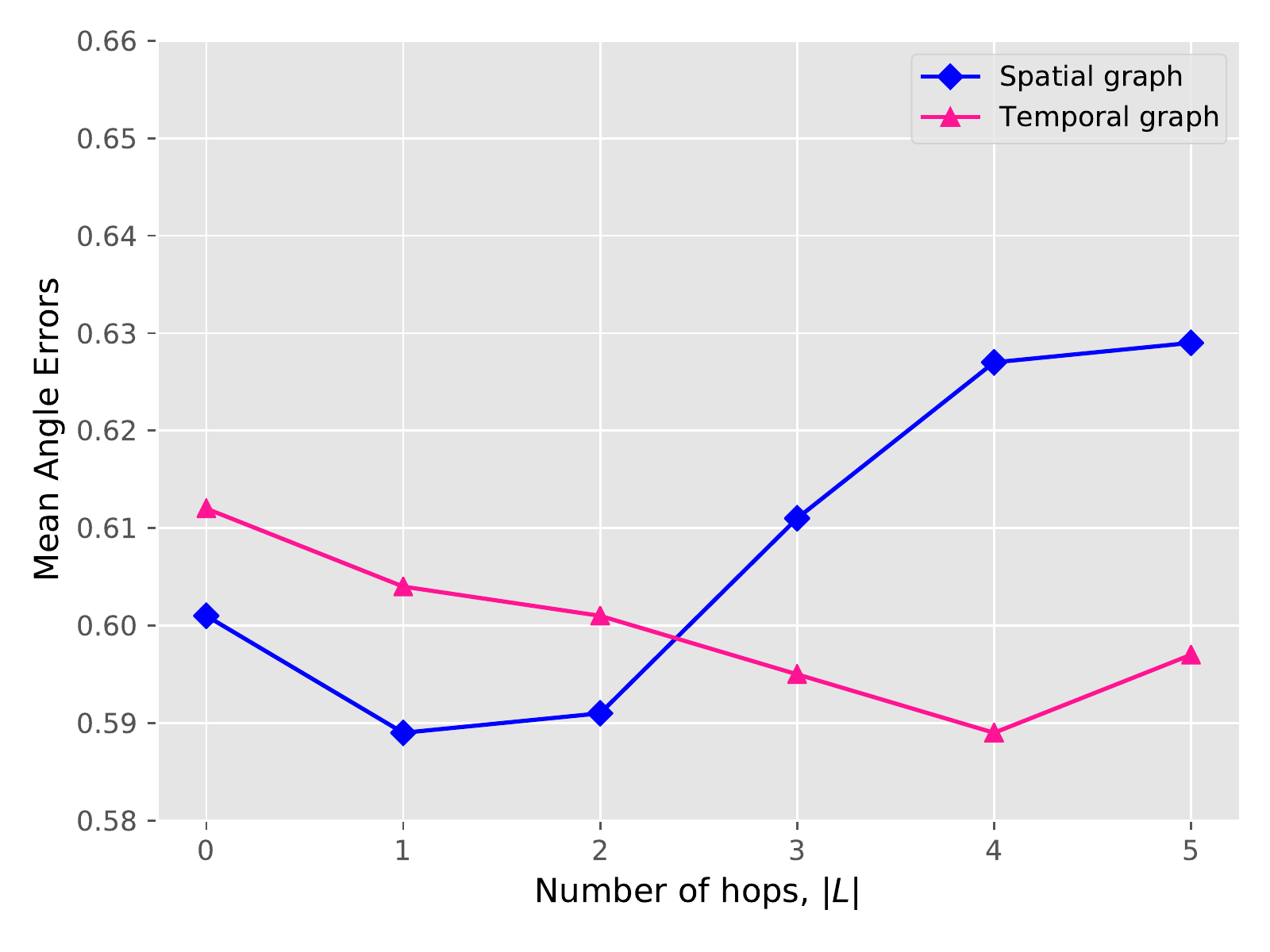} & 
     \includegraphics[width=0.47\columnwidth]{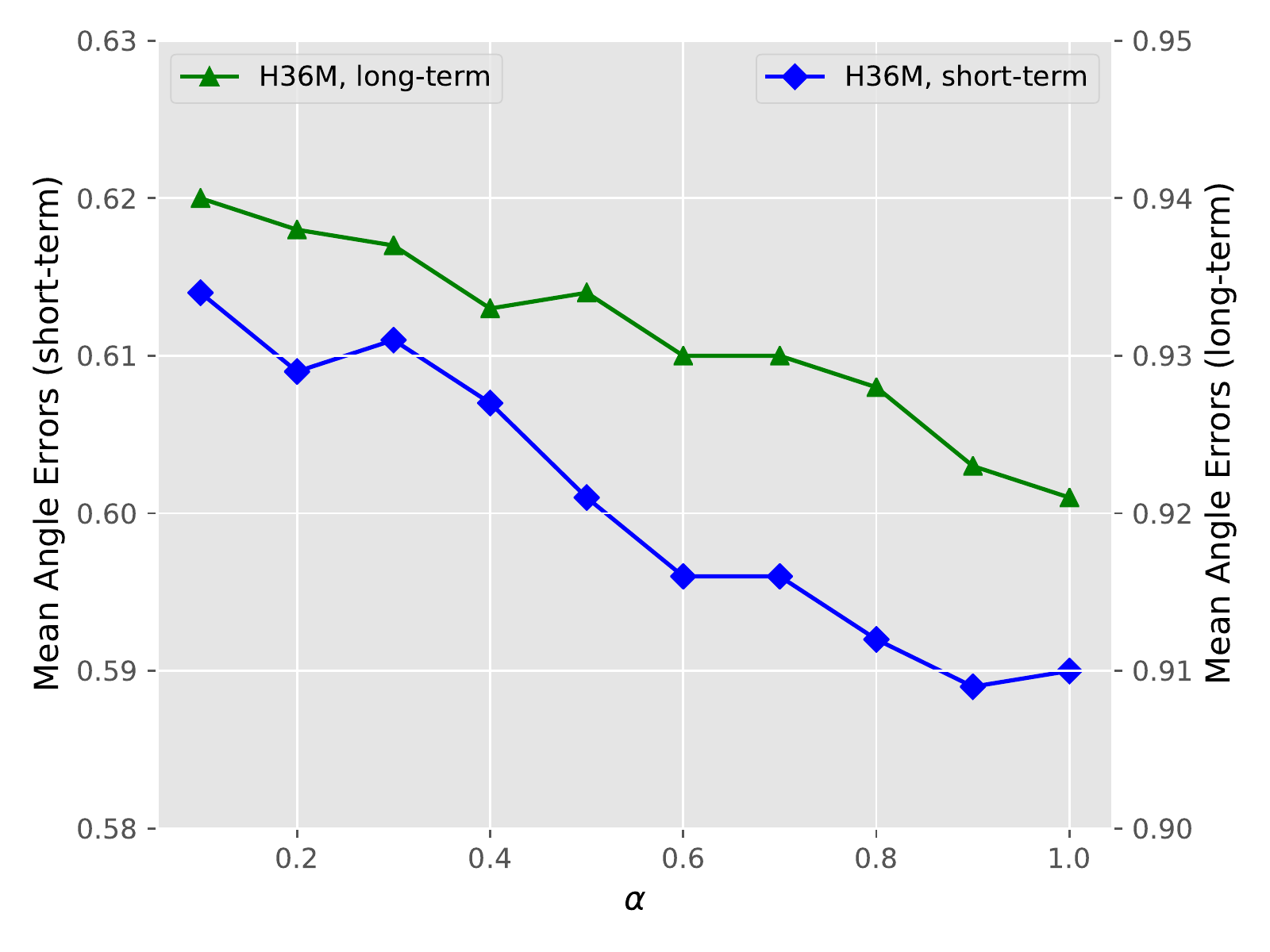} & 
     \includegraphics[width=0.47\columnwidth]{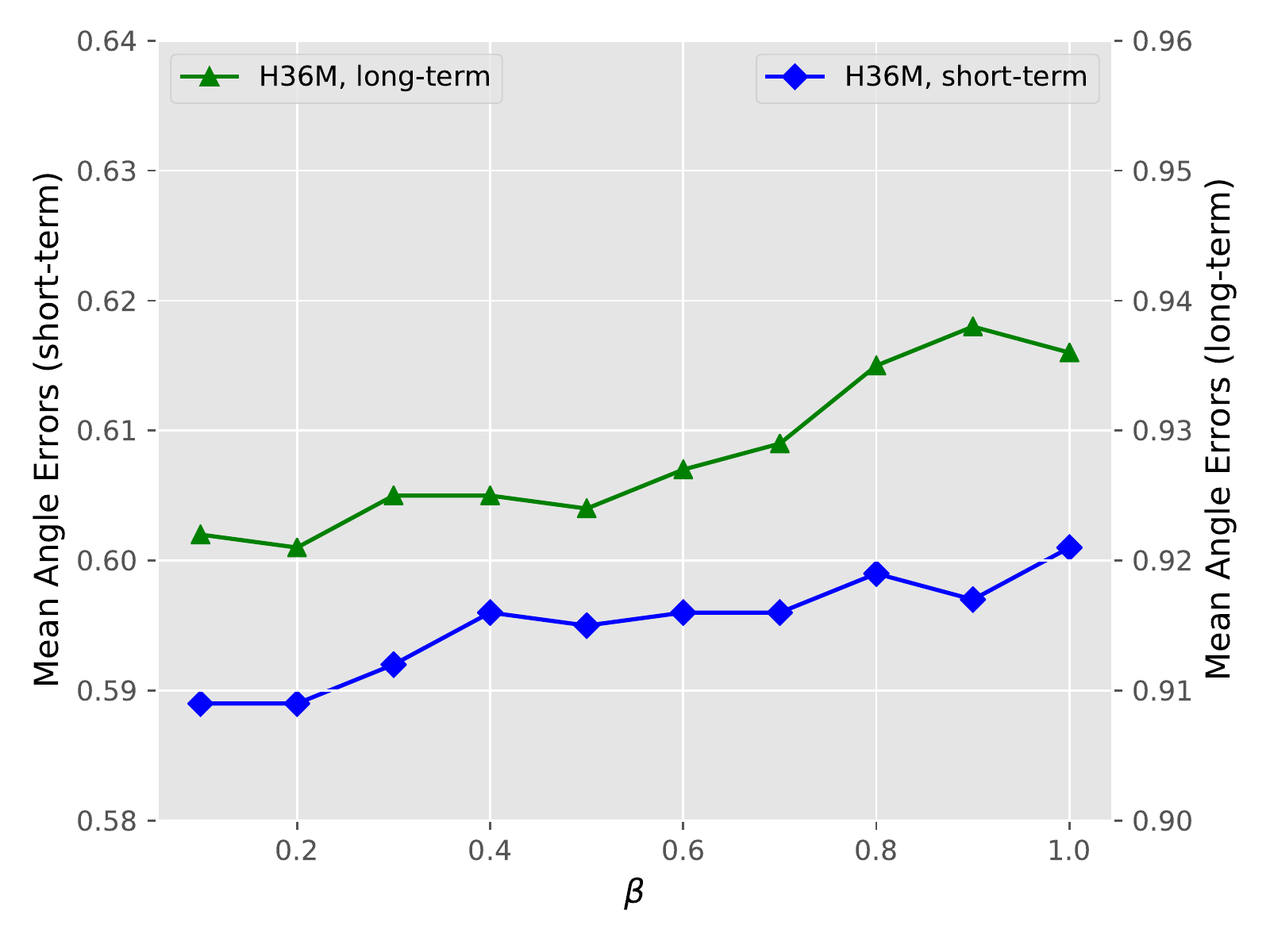} & 
     \includegraphics[width=0.47\columnwidth]{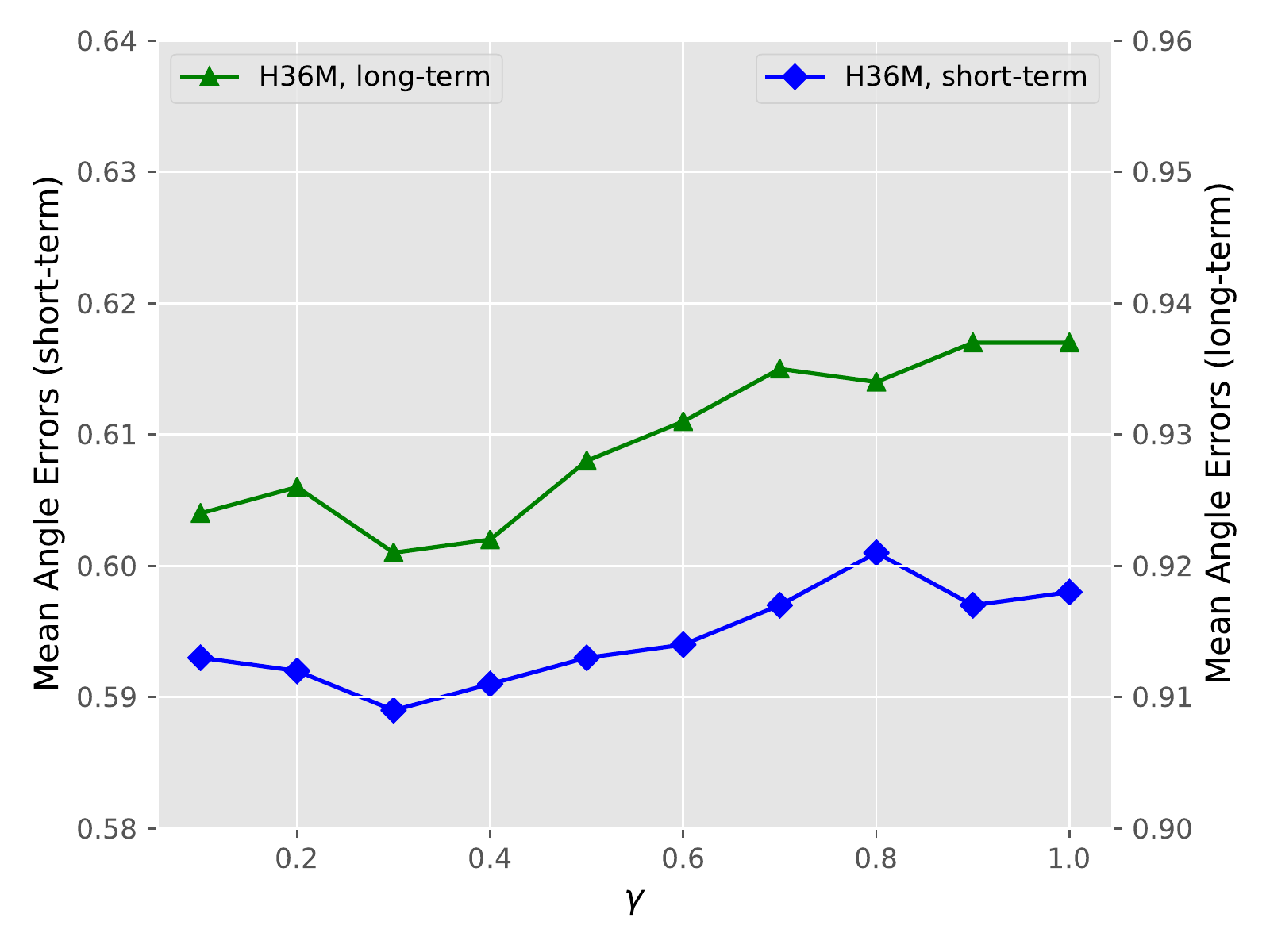} 
     \\
     {{\small (a) Effects of graph hops $L$.}} & 
     {{\small (b) Effects of $\alpha$ to prediction.}} & 
     {{\small (c) Effects of $\beta$ to prediction.}} & 
     {{\small (d) Effects of $\gamma$ to prediction.}}
  \end{tabular}
\end{center}
\vspace{-10pt}
\caption{\small The effects of thegraph hops in spatial and temporal graphs (a), as well as the coefficients of loss terms, $\alpha$ (b), $\beta$ (c) and $\gamma$ (d), to MAEs of motion prediction.}
\vspace{-10pt}
\label{fig:curve_fig}
\end{figure*}

\begin{table}[t]
    \centering
    \caption{MAEs of MST-GNN with various temporal scales (denoted by percentages of the compressed video lengths) for short-term prediction at 4 times and mean MAEs over all 400 ms.}
    \vspace{-2mm}
    \footnotesize
    \renewcommand{\arraystretch}{1.0}
    \resizebox{1\columnwidth}{!}{

        \begin{tabular}{|ccccc|ccccc|}
        \hline
        \multicolumn{5}{|c}{Temporal scale} & \multicolumn{5}{|c|}{MAEs} \\
        \hline
        $1$ & $1/2$ & $1/3$ & $1/4$ & $1/5$ & 80 & 160 & 320 & 400 & Avg.\\
        \hline
        \checkmark &  &  &  &  & 0.267 & 0.509 & 0.824 & 0.941 & 0.597 \\
        \checkmark & \checkmark &  &  &  & 0.265 & 0.510 & 0.818 & 0.935 & 0.595\\
        \checkmark &  & \checkmark &  &  & 0.265 & 0.506 & 0.821 & 0.935 & 0.594\\
        \checkmark &  &  & \checkmark &  & 0.265 & 0.501 & 0.817 & 0.931 & 0.593\\ 
        \checkmark &  &  &  & \checkmark & 0.267 & 0.502 & 0.818 & 0.933 & 0.594 \\ 
        \checkmark & \checkmark & \checkmark &  &  & {\bf 0.261} & {\bf 0.496} & {\bf 0.794} & {\bf 0.919} & {\bf 0.589} \\
        \checkmark & \checkmark &  & \checkmark &  & 0.263 & 0.501 & {0.814} & 0.922 & 0.591 \\
        \checkmark & \checkmark &  &  & \checkmark & {\bf 0.261} & {0.499} & 0.815 & {0.921} & {\bf 0.589} \\
        \checkmark & \checkmark & \checkmark & \checkmark &  & 0.263 & 0.501 & 0.816   & 0.925 & 0.592\\
        \checkmark & \checkmark & \checkmark &  & \checkmark & 0.263 & 0.499 & 0.816 & 0.920 & 0.591 \\
        \checkmark & \checkmark & \checkmark & \checkmark & \checkmark & 0.264 & 0.597 & 0.819 & 0.925 & 0.592 \\
        \hline
        \end{tabular}}
    \label{tab:TemporalScaleMAE}
    \vspace{-2mm}
\end{table}
For the multiscale temporal graphs, we test four abstract temporal scales that use $1/5$ to $1/2$ times of lengths of the original video. We present the MAEs of short-term prediction in Table~\ref{tab:TemporalScaleMAE}. When we use three temporal scales that use $100\%$, $50\%$ and $33\%$ video lengths, the MST-GNN achieves better performance. Several combinations of three temporal scales result in relatively stable results; fewer than three scales have no significant representation; and more than three scales cause sightly higher errors due to the heavy parameters.

\textbf{Effects of the number of MST-GCUs.}
\begin{table}[t]
    \centering
    \caption{MAEs and running times of MST-GNN with different numbers of MST-GCUs for short and long-term prediction on H3.6M.}
    \vspace{-2mm}
    \renewcommand{\arraystretch}{1.0}
    \resizebox{1\columnwidth}{!}{
        \begin{tabular}{|c|cccc|cc|cc|}
        \hline
         ~ & \multicolumn{6}{|c|}{MAE at different timestamps (ms)} &\multicolumn{2}{|c|}{running time (ms)}\\
        \hline
         MGCUs & 80&160&320&400&560&1000 & 400 & 1000\\
        \hline

        1 & 0.269 & 0.520 & 0.846 & 0.965 & 1.18 & 1.60 & 35.49 & 65.08 \\
        2 & 0.272 & 0.507 & 0.833 & 0.946 & 1.17 & 1.57 & 35.75 & 65.72 \\
        3 & 0.268 & 0.507 & 0.822 & 0.933 & 1.15 & 1.54 & 36.22 & 66.84 \\
        4 & {\bf 0.261} & {\bf 0.496} & {\bf 0.794} & {\bf 0.919} & {\bf 1.15} & {\bf 1.55} & 36.68 & 68.61 \\
        5 & 0.267 & 0.501 & 0.818 & 0.928 & {\bf 1.14} & 1.55 & 36.91 & 69.25 \\
        6 & 0.269 & 0.508 & 0.825 & 0.936 & 1.17 & 1.56 & 37.53 & 69.90 \\
        \hline
        \end{tabular}}
    \label{tab:layer}
    \vspace{-3mm}
\end{table}
To validate the effects of multiple MST-GCUs in the encoder, we tune the numbers of MST-GCUs from $1$ to $6$ and show the prediction errors and running time costs for short and long-term prediction on H3.6M, which are presented in Table~\ref{tab:layer}. We see that, when we adopt $1$ to $4$ MST-GCUs, the prediction MAEs fall and time costs rise continuously; when we use $5$ or $6$ MST-GCUs, the prediction errors are stably low, but the time costs rise higher. Therefore, we select to use $4$ MST-GCUs, resulting in precise prediction and high running efficiency.

\textbf{Effects of different spatio-temporal graph hops.}
Here we study the different numbers of graph hops in the spatial and temporal graph convolution, which are defined by the orders of graph adjacency matrices in~\eqref{eq:spatial_convolution} and~\eqref{eq:temporal_convolution}. We vary the graph hops $L$ for spatial and temporal graphs respectively to test the model on H3.6M. The average MAEs of motion prediction in both short-term and long-term are shown in Fig.~\ref{fig:curve_fig} (a). We see that, with the trainable spatial graphs that build edges between any distant vertices as implicit correlations, the model achieves the lowest prediction errors when the graph hop $L=1$; as for temporal graph, when the graph hop $L=4$, the model achieves the lowest errors, where a long-range receptive field along time is crucial for sequential information capturing.

\textbf{Effect of different terms in the loss.}
Here we investigate the effects of different loss terms proposed for model training, i.e. $\mathcal{L}_{\rm pred}$, $\mathcal{L}_{\rm gram}$ and $\mathcal{L}_{\rm ent}$; see~\eqref{eq:pred},~\eqref{eq:gram} and~\eqref{eq:ent}. We first test the different combinations of the three terms, where we select 1 to 3 of them to train the model. Note that when we use more than one loss terms, we choose the best coefficient corresponding to each term. The MAEs of the models trained by different loss functions are presented in Table~\ref{tab:loss_combine}.
\begin{table}[t]
    \centering
    \caption{MAEs of MST-GNN trained with the combinations of different loss terms for short-term motion prediction, as well as the average MAEs over 400 ms are presented.}
    \vspace{-2mm}
    \footnotesize
    \renewcommand{\arraystretch}{1.0}
    \resizebox{1\columnwidth}{!}{

        \begin{tabular}{|ccc|ccccc|}
        \hline
        \multicolumn{3}{|c|}{Loss terms} & \multicolumn{5}{c|}{MAEs} \\
        \hline
        $\mathcal{L}_{\rm pred}$ & $\mathcal{L}_{\rm gram}$ & $\mathcal{L}_{\rm ent}$ & 80 & 160 & 320 & 400 & Avg.\\
        \hline
        \checkmark &  &  &                      0.272 & 0.517 & 0.825 & 0.945 & 0.616\\
         & \checkmark &  &                      0.306 & 0.573 & 0.939 & 1.070 & 0.680\\
        \checkmark & \checkmark & &             0.263 & 0.498 & 0.816 & 0.923 & 0.591\\
        \checkmark &  & \checkmark &            0.265 & 0.498 & 0.819 & 0.926 & 0.592\\
         & \checkmark & \checkmark &            0.279 & 0.552 & 0.898 & 0.985 & 0.673\\
        \checkmark & \checkmark & \checkmark &  {\bf 0.261} & {\bf 0.494} & {\bf 0.814} & {\bf 0.919} & {\bf 0.589} \\
        \hline
        \end{tabular}}
    \label{tab:loss_combine}
    \vspace{-2mm}
\end{table}
We see that combining all three loss terms enables the proposed MST-GNN to obtain the best performance. Using one or two terms of loss has limited constraints for model training. Notably, regularizing the multiscale spatial graph generation, $\mathcal{L}_{\rm ent}$ also effectively improves the prediction accuracies.

We next test the effects of the loss coefficients for motion prediction. We vary the coefficient of each loss term, while the other coefficients are set as the most effective ones and kept unchanged. Fig.~\ref{fig:curve_fig} (b), (c), and (d) shows the MAE as a function of the varying coefficients on H3.6M.
We see that 1) for $\alpha$, which scales the $\ell_1$-based  prediction loss $\mathcal{L}_{\rm pred}$, a large $\alpha$ helps to generate more accurate motions close to the ground-truth; 2) for $\beta$ and $\gamma$, we need to select the appropriate values to balance the overall loss function and obtain lower MAEs, where we set $\beta=0.01$ and $\gamma=0.03$.

\vspace{-3mm}
\subsection{Analysis of Trainable Graph Pooling and Unpooling}

We first quantitatively evaluate the proposed data-driven multiscale graph construction module. We compare our method to several other graph downscaling methods on the same MST-GNN framework: 1) random averaging - `MST-GNN (random average)', where we randomly average joints to form specific numbers of clusters; 2) self-attention graph pooling~\cite{Lee_2019_ICML_SAG} - `MST-GNN (SAG-Pool)', where we apply graph convolution to learn an attention mask and preserve top-k joints with high attention scores; and 3) the predefined rules adopted from DMGNN~\cite{Li_2020_CVPR} - `MST-GNN (DMGNN-fix)'. We conduct short-term and long-term prediction with these methods on H3.6M, the prediction results are presented in Table~\ref{tab:simplepooling}.
\begin{table}[t]
      \centering
      \caption{The average prediction MAEs of four models with different multiscale spatial graph construction methods, for both short-term and long-term prediction on H3.6M.}
      \vspace{-2mm}
      \renewcommand{\arraystretch}{1.0}
      \resizebox{1\columnwidth}{!}{
      \begin{tabular}{|c|cccc|cc|}
          \hline
          ~ & \multicolumn{6}{|c|}{Average} \\
          \hline
          milisecond & 80 & 160 & 320 & 400 & 560 & 1000 \\
          \hline
          MST-GNN (random average) & 0.29 & 0.52 & 0.85 & 0.96 & 1.22 & 1.66 \\
          MST-GNN (DMGNN-fix) & 0.28 & 0.52 & 0.83 & 0.95 & 1.19 & 1.59 \\
          MST-GNN (SAG-Pool) & 0.27 & 0.53 & 0.83 & 0.94 & 1.20 & 1.62 \\
          MST-GNN & 0.26 & 0.50 & 0.79 & 0.92 & 1.15 & 1.55 \\
          \hline
      \end{tabular}}
      \vspace{-1mm}
      \label{tab:simplepooling}
  \end{table}
  We see that, i) based on the same MST-GNN framework, the model with the proposed data-driven multiscale graph construction consistently outperforms the previous human-prior-based multiscale graph construction in DMGNN, reflecting the advantage of module learning; and iii) our data-driven multiscale graph construction method significantly outperforms other data-driven graph pooling method, SAG-Pool, reflecting our nontrivial design.

\begin{figure}[t]
    \centering
    \includegraphics[width=0.8\columnwidth]{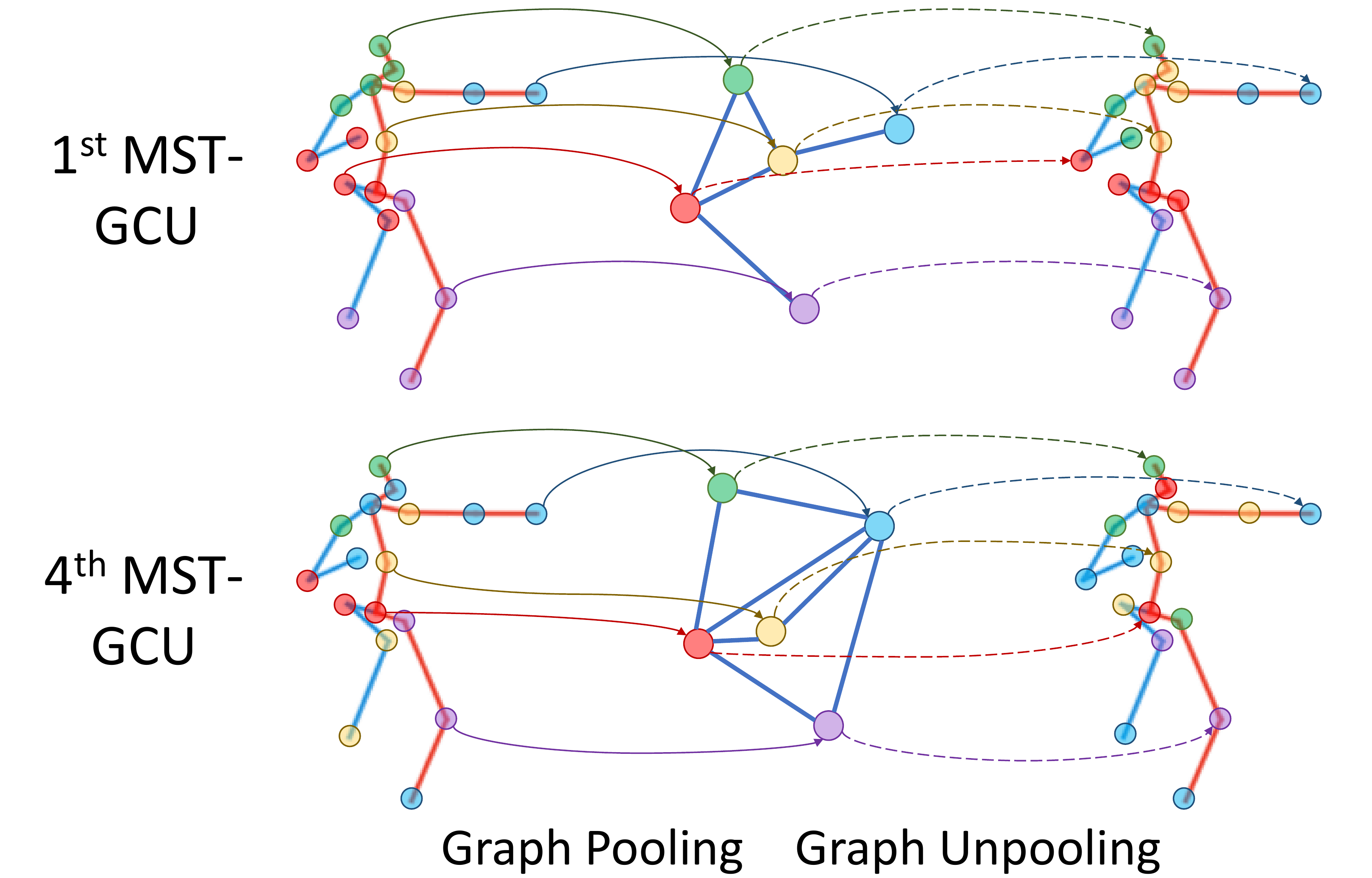}
    \caption{The trainable multiscale spatial graph pooling and unpooling at two MST-GCUs for the action of `Posing' in H3.6M.}
    \vspace{-5mm}
    \label{fig:PoolingViz}
\end{figure}

Furthermore, we investigate the properties of the proposed trainable graph pooling and unpooling. We mainly focus on the graph pooling and unpooling operations of multiscale spatial graphs, which are inferred based on the input motion data. The graph pooling and unpooling for the action of `Posing' in H3.6M in two MST-GCUs are illustrated in Fig.~\ref{fig:PoolingViz}. We show two scales of spatial graphs: the original joint-scale and the $2$nd scale containing $5$ vertices. We show the graph pooling and unpooling in the $1$st and $4$th MST-GCUs, which carries different spatial information in the model.
We see that, at the $1$st MST-GCU, the graph pooling operator tends to cluster some nearby joints due to the low-level features and small distances between joints; at the $4$th MST-GCU, the pooling operator capture some distant but collaborative moving joints which perform coordination during motion. As for the unpoolings at both the $1$st and $4$th MST-GCUs, they approximately map the abstract vertices to the positions corresponding to the graphs before pooling.

\vspace{-3mm}
\subsection{Analysis of Various Types of Temporal Graphs}
To evaluate the effectiveness of our temporal graph, we compare it to two other designs on the same MST-GNN framework. The first design is called MST-GNN (temporal context), where we construct the temporal graph $G_{\rm t} = (\mathcal{V}_{\rm t}, {\bf A}_{\rm t})$ as a bipartite graph between any two joints at consecutive poses, where $\mathcal{V}_{\rm t}$ denotes the node set that contains $2M$ nodes, and ${\bf A}_{\rm t} \in \mathbb{R}^{2M \times 2M}$ denotes the adjacency matrix. Given $G_{\rm t}$, we apply graph convolution on it. Note that  all the edge weights of the graph are adaptively tuned during the training process to model the flexible context. The second design is called MST-GNN (GCLNC-TG), which employs the temporal graph from~\cite{Zhong_2019_CVPR}, where the temporal graph is built based on the frame distances  (numbers of interval frames); that is, the edge weights between any two frames are obtained from the negative exponents of distances.

We compare the proposed MST-GNN to the two variants on H3.6M for both short-term and long-term motion prediction. The average MAE at each prediction step is presented in Table~\ref{tab:integratedgraph}.
  \begin{table}[t]
      \centering
      \caption{The comparison in terms of temporal graphs among three model variants: 1) MST-GNN with the proposed temporal graph (MST-GNN) and 2) MST-GNN with the temporal graph over incorresponding joints at different frames (MST-GNN (temporal context)), and MST-GNN with the temporal graph in GCLNC (MST-GNN (GCLNC-TG)), on H3.6M.}
      \vspace{-2mm}
      \renewcommand{\arraystretch}{1.0}
      \resizebox{1\columnwidth}{!}{
      \begin{tabular}{|c|cccc|cc|}
          \hline
          ~ & \multicolumn{6}{|c|}{Average} \\
          \hline
          milisecond & 80 & 160 & 320 & 400 & 560 & 1000 \\
          \hline
          MST-GNN & 0.26 & 0.50 & 0.79 & 0.92 & 1.15 & 1.55 \\
          MST-GNN (temporal context) & 0.28 & 0.53 & 0.86 & 0.96 & 1.24 & 1.77 \\
          MST-GNN (GCLNC-TG) & 0.29 & 0.53 & 0.84 & 0.95 & 1.19 & 1.61 \\
          \hline
      \end{tabular}}
      \vspace{-5mm}
      \label{tab:integratedgraph}
  \end{table}
  We see that, the proposed MST-GNN achieves much lower MAEs by $8.15\%$ and $4.12\%$ in average than MST-GNN (temporal context) and MST-GNN (GCLNC-TG), respectively. The comparison between MST-GNN and MST-GNN (temporal context) reflects the two-fold properties. First, forcibly modeling the temporal relations between the incorresponding joints might show information redundancy; that is, due to the iterative spatial and temporal information propagation, the contextual relations between incorresponding joints could be indirectly build through the spatio-temporal graphs, in other words, the information could be propagated between incorresponding joints at different frame through the combined spatial and temporal graphs together. Second, the fully trainable temporal graph that considers temporal contexts would introduce much freedom of parameters, causing the difficulty of training. The comparison between MST-GNN and MST-GNN (GCLNC-TG) also shows that, the effectiveness of the proposed temporal graph is mainly obtained by flexible edge weights to describe highly flexible temporal correlations, while GCLNC determines the temporal effects only accordingto  the  temporal  distances.  In  fact,  the  strong  temporal  correlations  might  not  be  built  soly  based-on temporal distances; it could also be related to historical states and long-term collaboration.

\section{Conclusions}
\label{sec:Conclusion}
We construct multiscale spatio-temporal graphs to represent a human body performing different motions and propose multiscale spatio-temporal multiscale graph neural networks (MST-GNN) with an encoder-decoder framework for 3D skeleton-based human motion prediction. In the encoder, We develop a series of multiscale spatio-temporal graph computational units (MST-GCU) to extract features; in the decoder, we develop a graph-based attention GRU (GA-GRU) for pose generation. The results show that the proposed model outperforms most state-of-the-art methods for both short and long-term human motion prediction.

\section*{Acknowledgement}
\vspace{-1mm}
This work is supported by the National Key Research and Development Program of China (2019YFB1804304), SHEITC (2018-RGZN-02046), 111 plan (BP0719010), Shanghai "Science and Technology Innovation Plan" Key Research Program of Artificial Intelligence (21511100900), STCSM (18DZ2270700), State Key Laboratory of UHD Video and Audio Production and Presentation.

\ifCLASSOPTIONcaptionsoff
  \newpage
\fi



%

\bibliographystyle{IEEEtran}
\bibliography{IEEEabrv}

\vspace{-10mm}
\begin{IEEEbiography}
[{\includegraphics[width=1in,height=1.25in,clip,keepaspectratio]{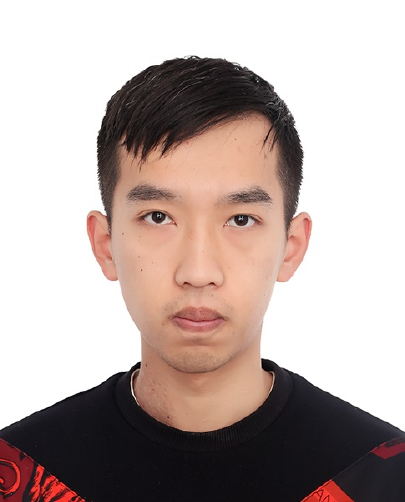}}]
{Maosen Li} recieved the B.E. degree in optical engineering from University of Electronic Science and Technology of China (UESTC), Chengdu, China, in 2017. He is working toward the Ph.D. degree at Cooperative Meidianet Innovation Center in Shanghai Jiao Tong University since 2017. His research interests include computer vision, machine learning, graph representation learning, and video analysis. He is the reviewer of some prestigious international journals and conferences, including IEEE-TPAMI, IEEE-TNNLS, IJCV, IEEE-TMM, PR, ICML, NeurIPS and AAAI. He is a student member of the IEEE.
\end{IEEEbiography}
\vspace{-10mm}
\begin{IEEEbiography}
[{\includegraphics[width=1in,height=1.25in,clip,keepaspectratio]{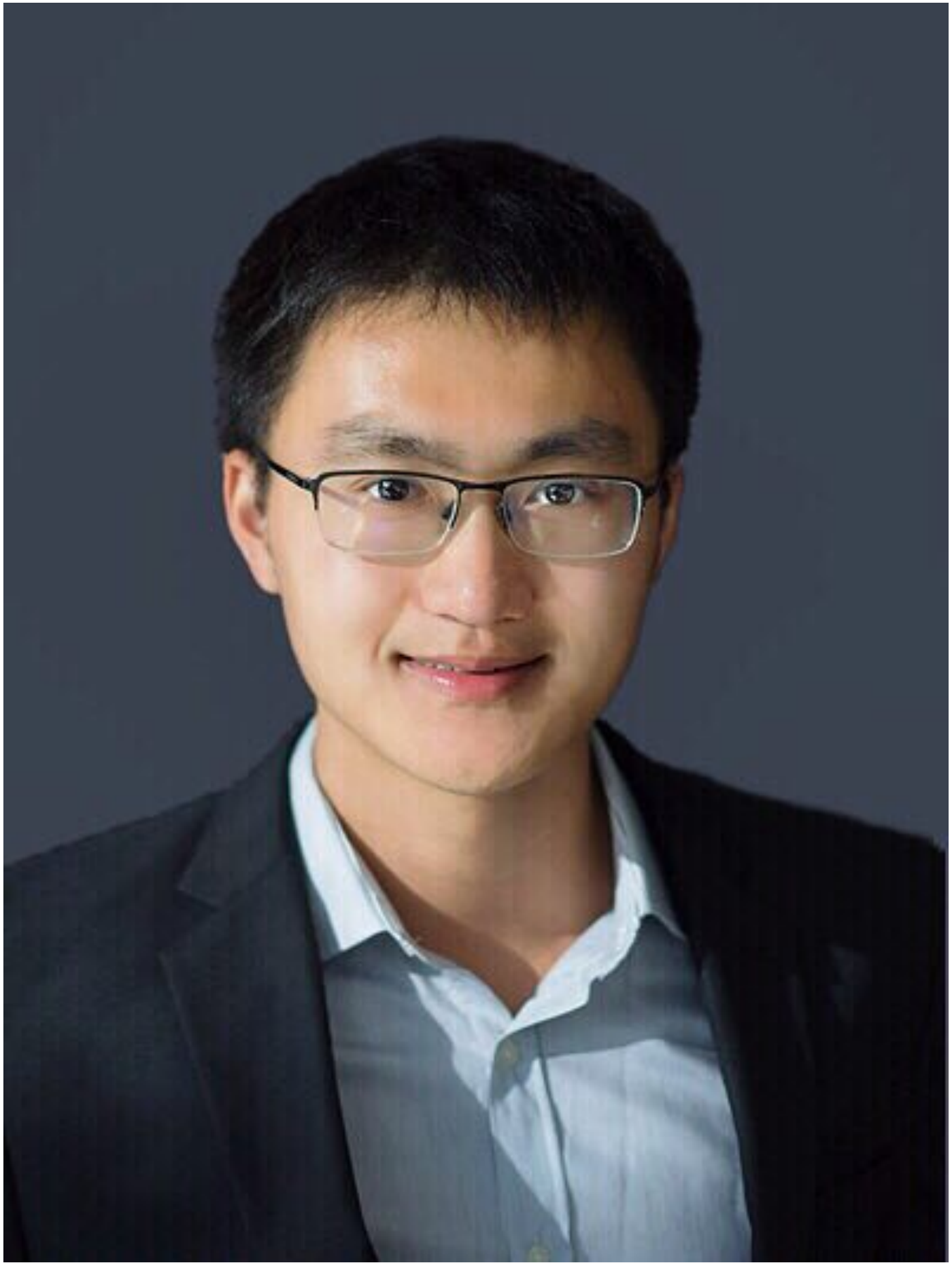}}]
{Siheng Chen} is an associate professor at Shanghai Jiao Tong University. Before that, he was a research scientist at Mitsubishi Electric Research Laboratories (MERL). Before joining MERL, he was an autonomy engineer at Uber Advanced Technologies Group, working on the perception and prediction systems of self-driving cars. Before joining Uber, he was a postdoctoral research associate at Carnegie Mellon University. Chen received the doctorate in Electrical and Computer Engineering from Carnegie Mellon University in 2016, where he also received two masters degrees in Electrical and Computer Engineering and Machine Learning, respectively. He received his bachelor's degree in Electronics Engineering in 2011 from Beijing Institute of Technology, China. Chen was the recipient of the 2018 IEEE Signal Processing Society Young Author Best Paper Award. His coauthored paper received the Best Student Paper Award at IEEE GlobalSIP 2018. He organized the special session "Bridging graph signal processing and graph neural networks" at ICASSP 2020. His research interests include graph signal processing, graph neural networks and autonomous driving. He is a member of IEEE.
\end{IEEEbiography}
\vspace{-10mm}
\begin{IEEEbiography}
[{\includegraphics[width=1in,height=1.25in,clip,keepaspectratio]{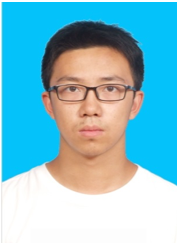}}]
{Yangheng Zhao} received his B.E. degree from in information engineering from Shanghai Jiao Tong Univeristy, Shanghai, China, in 2020. He is currently studying as a master student at Cooperative Meidianet Innovation Center in Shanghai Jiao Tong University. His research interests include computer vision, graph representation learning, 3D generalization and graphics.
\end{IEEEbiography}
\vspace{-10mm}
\begin{IEEEbiography}
[{\includegraphics[width=1in,height=1.25in,clip,keepaspectratio]{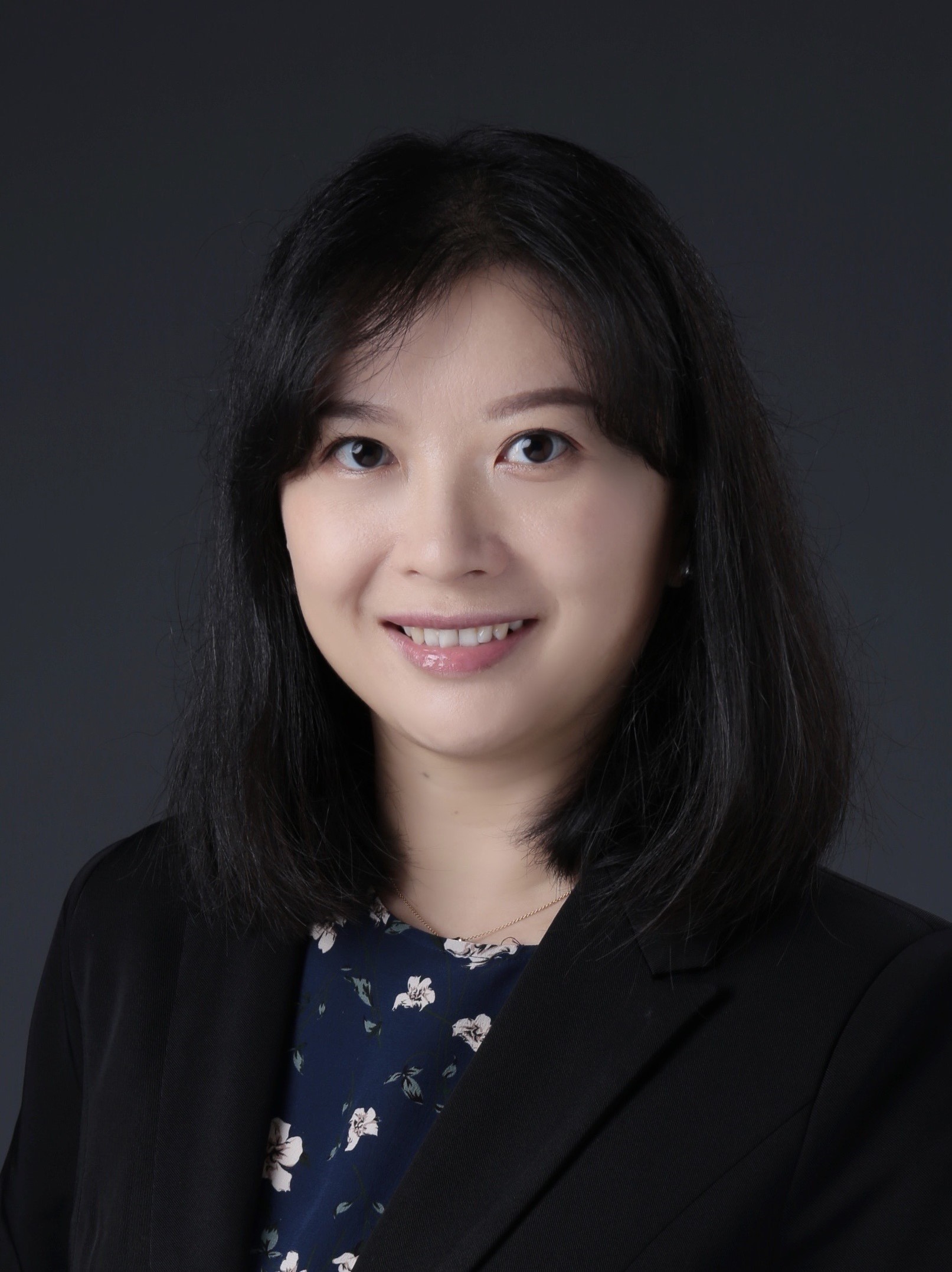}}]
{Ya Zhang}
is currently a professor at the Cooperative Medianet Innovation Center, Shanghai Jiao Tong University. Her research interest is mainly in machine learning with applications to multimedia and healthcare. Dr. Zhang holds a Ph.D. degree in Information Sciences and Technology from Pennsylvania State University and a bachelor's degree from Tsinghua University in China. Before joining Shanghai Jiao Tong University, Dr. Zhang was a research manager at Yahoo! Labs, where she led an R\&D team of researchers with strong backgrounds in data mining and machine learning to improve the web search quality of Yahoo international markets. Prior to joining Yahoo, Dr. Zhang was an assistant professor at the University of Kansas with a research focus on machine learning applications in bioinformatics and information retrieval. Dr. Zhang has published more than 70 refereed papers in prestigious international conferences and journals, including TPAMI, TIP, TNNLS, ICDM, CVPR, ICCV, ECCV, and ECML. She currently holds 5 US patents and 4 Chinese patents and has 9 pending patents in the areas of multimedia analysis. She was appointed the Chief Expert for the 'Research of Key Technologies and Demonstration for Digital Media Self-organizing' project under the 863 program by the Ministry of Science and Technology of China. She is a member of IEEE.
\end{IEEEbiography}
\vspace{-10mm}
\begin{IEEEbiography}
[{\includegraphics[width=1in,height=1.25in,clip,keepaspectratio]{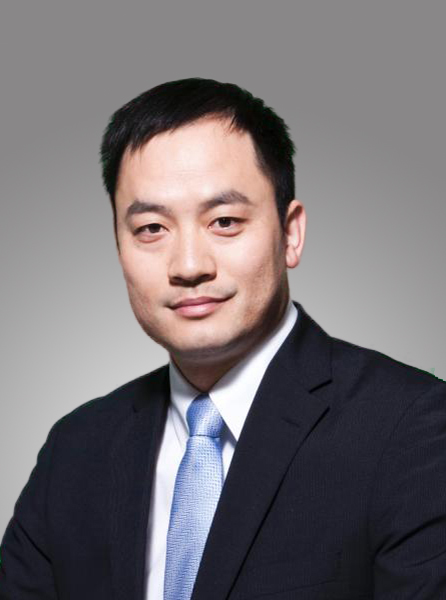}}]
{Yanfeng Wang} received the B.E. degree in information engineering from the University of PLA, Beijing, China, and the M.S. and Ph.D. degrees in business management from the Antai College of Economics and Management, Shanghai Jiao Tong University, Shanghai, China. He is currently the Vice Director of Cooperative Medianet Innovation Center and also the Vice Dean of the School of Electrical and Information Engineering with Shanghai Jiao Tong University. His research interest mainly include media big data and emerging commercial applications of information technology.
\end{IEEEbiography}
\vspace{-10mm}
\begin{IEEEbiography}
[{\includegraphics[width=1in,height=1.25in,clip,keepaspectratio]{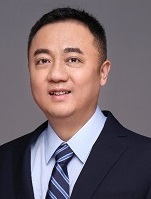}}] 
{Qi Tian} is currently the Chief Scientist of computer vision at Huawei Cloud \& AI and a full professor with the Department of Computer Science at the University of Texas at San Antonio (UTSA). He was a tenured associate professor during 2008-2012 and a tenure-track assistant professor during 2002-2008. From 2008 to 2009, he took faculty leave for one year at Microsoft Research Asia (MSRA) as Lead Researcher in the Media Computing Group. Dr. Tian received his Ph.D. in ECE from the University of Illinois at Urbana-Champaign (UIUC) in 2002 and received his B.E. degree in electronic engineering from Tsinghua University in 1992 and his M.S. degree in ECE from Drexel University in 1996. Dr. Tian's research interests include multimedia information retrieval, computer vision, pattern recognition. He has published over 700 refereed journal and conference papers. He was the coauthor of a Best Paper at ACM ICMR 2015, a Best Paper at PCM 2013, a Best Paper at MMM 2013, a Best Paper at ACM ICIMCS 2012, a Top 10\% Paper at MMSP 2011, a Best Student Paper at ICASSP 2006, a Best Student Paper Candidate at ICME 2015, and a Best Paper Candidate at PCM 2007. Dr. Tian received the 2017 UTSA President's Distinguished Award for Research Achievement; the 2016 UTSA Innovation Award; the 2014 Research Achievement Awards from the College of Science, UTSA; the 2010 Google Faculty Award; and the 2010 ACM Service Award. He is an associate editor of many journals and on the Editorial Board of the Journal of Multimedia (JMM) and Journal of Machine Vision and Applications (MVA). He is a fellow of the IEEE.
\end{IEEEbiography}

\end{document}